\documentclass{article}

\usepackage[final]{cpal_2025}

\usepackage{amsmath,amsfonts,bm}

\def\Figref#1{Figure~\ref{#1}}

\def\eqref#1{equation~\ref{#1}}
\def\Eqref#1{Equation~\ref{#1}}

\def\Algref#1{Algorithm~\ref{#1}}

\def\Tabref#1{Table~\ref{#1}}

\def\1{\bm{1}}

\DeclareMathAlphabet{\mathsfit}{\encodingdefault}{\sfdefault}{m}{sl}
\SetMathAlphabet{\mathsfit}{bold}{\encodingdefault}{\sfdefault}{bx}{n}

\DeclareMathOperator*{\argmin}{arg\,min}

\usepackage{hyperref} 
\usepackage{url}

\usepackage{graphicx}
\usepackage{booktabs}
\usepackage{amsfonts}
\usepackage{nicefrac}
\usepackage{microtype}
\usepackage{xcolor}

\usepackage{amsmath}
\usepackage{amssymb}
\usepackage{mathtools}

\usepackage{algorithmicx}
\usepackage{algorithm}
\usepackage{algpseudocode}

\usepackage{booktabs, caption, makecell}
\usepackage{threeparttable}
\usepackage{progressbar}
\usepackage{wrapfig}

\usepackage{multirow}
\usepackage{xcolor,colortbl}
\usepackage{siunitx} 

\usepackage{subcaption}

\definecolor{lightred}{rgb}{1,0.8,0.8}
\definecolor{lightgreen}{rgb}{0.8,1,0.8}
\definecolor{lightblue}{rgb}{0.88,0.96,1} 
\definecolor{lightgray}{rgb}{0.9,0.9,0.9}

\usepackage{comment}

\usepackage{tocloft}

\title{FedPeWS: Personalized Warmup via Subnetworks for Enhanced Heterogeneous Federated Learning}

\author{%
  Nurbek Tastan\textsuperscript{1}, ~Samuel Horv\'{a}th\textsuperscript{1}, ~Martin Tak\'{a}\v{c}\textsuperscript{1}, ~Karthik Nandakumar\textsuperscript{1,2} \\ 
  \textsuperscript{1}Mohamed bin Zayed University of Artificial Intelligence (MBZUAI), UAE \\ 
  \textsuperscript{2}Michigan State University (MSU), USA \\ 
  \texttt{\{nurbek.tastan,samuel.horvath,martin.takac\}@mbzuai.ac.ae, nandakum@msu.edu} 
}

\begin{document}

\maketitle

\begin{abstract}
  Statistical data heterogeneity is a significant barrier to convergence in federated learning (FL). While prior work has advanced heterogeneous FL through better optimization objectives, these methods fall short when there is \textit{extreme} data heterogeneity among collaborating participants. We hypothesize that convergence under extreme data heterogeneity is primarily hindered due to the aggregation of conflicting updates from the participants in the initial collaboration rounds. To overcome this problem, we propose a warmup phase where each participant learns a personalized mask and updates only a subnetwork of the full model. This \textit{personalized warmup} allows the participants to focus initially on learning specific \textit{subnetworks} tailored to the heterogeneity of their data. After the warmup phase, the participants revert to standard federated optimization, where all parameters are communicated. We empirically demonstrate that the proposed personalized warmup via subnetworks (\texttt{FedPeWS}) approach improves accuracy and convergence speed over standard federated optimization methods. The code can be found at \href{https://github.com/tnurbek/fedpews}{https://github.com/tnurbek/fedpews}. 
\end{abstract}

\section{Introduction} 
\label{section: intro} 

Federated learning (FL) is a distributed learning paradigm where participants collaboratively train a global model by performing local training on their data and periodically sharing local updates with the server. The server, in turn, aggregates the local updates to obtain the global model, which is then transmitted to the participants for the next round of training \citep{mcmahan2017communication}. While FL preserves data confidentiality by avoiding collating participant data at the server, \textit{statistical heterogeneity} between local data distributions is a significant challenge in FL \citep{kairouz2021advances}. Several attempts have been made to tackle heterogeneity via federated optimization algorithms \citep{wang2019matcha, khaled2019first, li2019convergence, li2020federated, karimireddy2020scaffold,tupitsa2024federated,sadiev2022decentralized,beznosikov2021decentralized}, dropout \citep{horvath2021fjord, alam2022fedrolex}, and batch normalization \citep{li2021fedbn}. 

Consider the scenario where multiple hospitals collaborate to learn a medical image classification model that works across imaging modalities and organs, where the data from each hospital pertains to a different modality (e.g., histopathology, CT, X-ray, ultrasound, etc.) and/or organ (e.g., brain, kidney, colon, etc.). Most of the existing heterogeneous FL algorithms fail when there is such \textit{extreme} data heterogeneity among collaborating participants. The primary reason for this failure is the high degree of conflicts between the local updates and the inherent bias the algorithms introduce, often favoring one modality (dataset) over the others. While enforcing a strong regularization constraint on the local updates \cite{li2020federated} can partially alleviate this problem, it dramatically slows down local learning and hence, convergence speed.

In this work, we explore an alternate approach to minimize the initial conflicts between heterogeneous participants by allowing participants in FL to initially train a partial subnetwork using only their local datasets. This warmup phase enables the participants to focus first on learning their local data well before engaging in broader collaboration. Thus, our proposed approach can be summarized as follows (see \Figref{fig: main-figure}). Initially, each participant uses a personalized binary mask, allowing them to first learn their local data distributions and optimize their local (sparse) models. During this warmup phase, participants transmit only their masked updates to the server, and this process continues for a certain number of collaboration rounds. At the end of the warmup phase, the participants switch to standard federated optimization for subsequent collaboration. Our contributions are as follows:

\begin{figure}[t!]
    \centering
    \includegraphics[width=\linewidth]{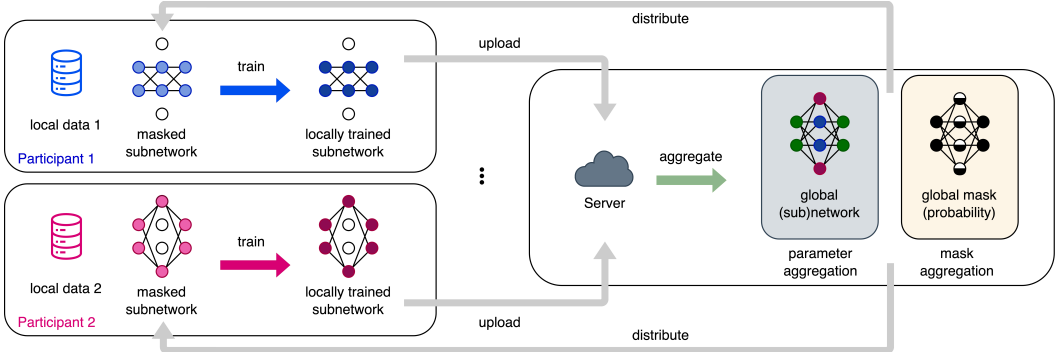} 
    \caption{Conceptual illustration of training personalized subnetworks in federated learning.}
    \label{fig: main-figure}
\end{figure}

\begin{enumerate}
    \item We introduce a novel concept in federated learning, termed as \textit{personalized warmup via subnetworks} (\texttt{FedPeWS}), which helps the global model to generalize to a better solution in fewer communication rounds. This is achieved through a neuron-level personalized masking strategy that is compatible with other FL optimization methods. 
    \item We propose an algorithm to \textit{identify suitable subnetworks} (subset of neurons) for each participant by simultaneously learning the personalized masks and parameter updates. The proposed algorithm does not make any assumptions regarding the data distributions and incorporates a \textit{mask diversity loss} to improve the coverage of all neurons in the global model. 
    \item For simple cases involving a small number of participants with known data distributions, we show that it is possible to skip the mask learning step and use fixed masks (that partition the network) determined by the server. We refer to this variant as \texttt{FedPeWS-Fixed}. 
    \item We empirically demonstrate the efficacy of the \texttt{FedPeWS} approach under both extreme non-i.i.d. and i.i.d. data scenarios using three datasets: a custom synthetic dataset, a combination of MNIST and CIFAR-10 datasets, a combination of three distinct medical datasets (PathMNIST, OCTMNIST and TissueMNIST), and CIFAR-100 dataset. 
\end{enumerate}

\section{Related Work} 

\paragraph{Collaborative Learning.} FL is a distributed learning paradigm that addresses data confidentiality concerns \citep{kairouz2021advances}, particularly in environments where data can not be centralized due to regulatory or practical reasons \citep{albrecht2016gdpr}. One of the seminal FL algorithms, FedAvg \citep{mcmahan2017communication}, involves participants training models locally on their data and periodically transmitting their model parameters to a central server. The server averages these parameters to update the global model, which is then redistributed to the participants for further local refinement. FedAvg has inspired a plethora of variants and extensions aimed at enhancing performance \citep{karimireddy2020scaffold, li2020federated, mishchenko2022proxskip}, scalability \citep{guo2023federated, al-shedivat2021federated}, communication efficiency \citep{ullah2023private, rahimi2023evofed, isik2022sparse}, privacy/confidentiality \citep{tastan2023capride, choquette2021capc, ullah2023private}, robustness \citep{li2019rsa}, and fairness \citep{xu2021gradient, jiang2023fair, tastan2024redefining}. For example, strategies such as weighted averaging or adaptive aggregation have been proposed to accommodate the non-i.i.d. nature of distributed data sources -- a scenario where data is not identically distributed across all participants, which can significantly hinder model performance \citep{li2020federated, wang2020tackling, karimireddy2020scaffold, li2021fedbn, wang2020federated, chezhegov2024local}. Specifically, FedProx \citep{li2020federated} addresses data heterogeneity by integrating a proximal term into the FedAvg framework. There is also a body of work that focuses on addressing the heterogeneity problem through personalization-based approaches, utilizing local-centric objectives \citep{gasanov22flix, hanzely2023personalized, yoon2021fedweit,sadiev2022decentralized,borodich2021decentralized, li2021ditto}. 

\paragraph{Independent Subnet Training.} Independent subnet training (IST) is a variant of distributed learning that focuses on enhancing model personalization and reducing communication overhead by training separate subnetworks for different participants \citep{yuan2022distributed}. IST distributes neurons of a fully connected neural network disjointly across different participants, forming a group of subnets. Then, each of these subnets is trained independently for one or more local SGD steps before synchronization. In every round, after broadcasting the server weights, each participant gets updated neurons to focus on, and the local subnet training continues. This approach led to different works along the line of using subnetwork training for efficiency \citep{horvath2021fjord, jiang2022prunefl, diao2020heterofl, nader2020feddropoutadaptive, alam2022fedrolex, li2021fedmask, mozaffari2021frl} in FL. In our work, we adopt IST's core principle of selecting neurons rather than focusing on weight values, which in turn narrows the search space. A key distinction between our method and IST lies in how the neurons are selected and the necessity of covering all neurons. Whilst IST typically involves random sampling of masks in each training round by the server and full coverage of neurons, we do not randomly sample neurons; instead, we use a learnable mask for each participant that is trained along with the parameters, and we relax the assumption of full coverage of neurons.

\begin{table}[t]
    \centering
    \caption{Comparison of approaches for handling data heterogeneity in federated learning.} 
    \resizebox{0.9\linewidth}{!}{ 
        \begin{tabular}{lcccc}
            \toprule
             & \multirow{2}{*}{\parbox{2.3cm}{\centering \textbf{Shared \\ Global Model}}} & \multirow{2}{*}{\parbox{2.5cm}{\centering \textbf{Level of Mask \\ Personalization}}} & \multirow{2}{*}{\parbox{2cm}{\centering {\textbf{Learnable \\ Mask}}}} & \multirow{2}{*}{\parbox{2cm}{\centering {\textbf{Learnable \\ Parameters}}}} \\ 
             & & & & \\ \midrule 
             FedPM \citep{isik2022sparse} & \checkmark & parameter-level & \checkmark & $\bm{\times}$ \\ 
             IST \citep{yuan2022distributed} & \checkmark & neuron-level & $\bm{\times}$ (random) & \checkmark \\ 
             LTH \citep{frankle2018lottery} & $\bm{\times}$ & parameter-level & $\bm{\times}$ (pruned) & \checkmark \\ 
             FedWeIT \citep{yoon2021fedweit} & $\bm{\times}$ & parameter-level & \checkmark & \checkmark \\ 
             FjORD \citep{horvath2021fjord} & \checkmark & parameter-level & $\bm{\times}$ (slimmed) & \checkmark \\ \midrule 
             \rowcolor{lightblue} \texttt{FedPeWS} (ours) & \checkmark & neuron-level & \checkmark & \checkmark \\ \bottomrule 
        \end{tabular}
    }
    \label{tab: methods-comparison}
\end{table}

\paragraph{Finding Subnetworks in FL.} Another relevant idea is the Lottery Ticket Hypothesis (LTH) \citep{frankle2018lottery}, which attempts to identify subnetworks within a larger network. LTH is a model personalization technique, which focuses on sparsifying the network to create a smaller-scale version that improves per-round communication efficiency. In contrast to LTH, our method is directed towards training a shared global model and simultaneously improving convergence speed (reducing the number of communication rounds). After LTH, there has been a growing interest in finding sparse and trainable networks at initialization \citep{mellor2021naswot, ji2021dynamic, li2020lotteryfl}. 
In \citep{li2020lotteryfl, huang2022achieving}, the goal is to learn a personalized model for each client with better communication efficiency. While the sparse subnetworks are learned through pruning in \citep{li2020lotteryfl}, dynamic sparse training-based mask search is employed in \citep{huang2022achieving}. The work by \citet{babakniya2022federated} focuses on model compression, i.e., learning a sparse server model by identifying a sparse subnetwork within the global model learned via FL. 
To elaborate further, \citep{li2020lotteryfl} leverages the lottery ticket hypothesis to find lottery ticket networks inside the main network, and only these networks are communicated during training. The subnetworks are located using pruning techniques, with pruning rates assigned by the server based on hard-coded thresholds for accuracy gains and target pruning rates. At the end of the collaboration, clients obtain personalized models. \citep{huang2022achieving} is a personalized FL technique that employs sparse-to-sparse training with the goal of reducing communication, computation, and memory costs. Sparse models are achieved by applying weight magnitude pruning combined with gradient information computed from a randomly sampled batch. Furthermore, \citep{babakniya2022federated} is a sparse learning framework primarily focused on model compression. It identifies subnetworks through sensitivity analysis on a randomly generated mask, which is used to identify a good initial mask for each layer. 
Our approach fundamentally differs from these methods by aiming to train a single global (dense) model while simultaneously improving the convergence speed (reducing the number of communication rounds required). \texttt{FedPeWS} identifies sparse, personalized subnetworks only during the warmup phase of the training to enable quicker/better convergence under extreme non-IID settings.  

Recently, in \citep{isik2022sparse}, sparse networks were found inside the main model to increase communication efficiency in FL. The proposed FedPM method focuses on finding a subnetwork by freezing the model weights and training for masks on a weight level, in contrast to IST, which works on a neuron level. FedPM utilizes the sigmoid function to obtain probability values from unbounded mask scores and then uses Bernoulli sampling to obtain binary masks. We use a similar approach in our \texttt{FedPeWS} algorithm to learn the neuron-level personalized masks.

Table \ref{tab: methods-comparison} presents a comprehensive comparison of our proposed method, \texttt{FedPeWS}, against existing literature for addressing data heterogeneity in federated learning. The comparison highlights key attributes of each method, including the level of mask personalization and whether the masks and parameters are learnable. 

\paragraph{Mixture of Experts.} Mixture of Experts (MoE) \citep{jacobs1991adaptive, jordan1994hierarchical} is a technique that trains multiple specialized models (experts) alongside a gating network that dynamically selects which experts to activate for a given input during both training and inference. Although MoE and our proposed method share the concept of leveraging different model components for learning, there is a fundamental distinction. \texttt{FedPeWS} is not designed to train multiple expert models; rather, it learns a single global model while identifying sparse, personalized subnetworks exclusively during the warmup phase of training. This warmup strategy serves as a crucial initialization step to facilitate efficient federated optimization, improving both convergence speed and generalization in heterogeneous FL environments.

\section{Preliminaries}
Our goal is to minimize a sum-structured federated learning optimization objective: 
\begin{equation}
    \begin{split}
        x^{\star} \leftarrow  \argmin_{x \in \mathbb{R}^d} \Bigg[ f(x) \coloneqq \frac{1}{N} \sum_{i=1}^N f_i(x) \Bigg], 
        \label{eq: optimization-problem}
    \end{split}
\end{equation}
where the components $f_i: \mathbb{R}^d \rightarrow \mathbb{R}$ are distributed among $N$ local participants and are expressed in a stochastic format as $f_i(x) \coloneqq \mathbb{E}_{\xi \sim \mathcal{D}_i}\big[F_i(x, \xi)\big]$. Here, $\mathcal{D}_i$ represents the distribution of $\xi$ at participant $i \in [N] \coloneqq \{1,\ldots,N\}$. This problem encapsulates standard empirical risk minimization as a particular case when each $\mathcal{D}_i$ is represented by a finite set of $n_i$ elements, i.e., $\xi_i = \left\{\xi_i^1, \ldots, \xi_i^{n_i} \right\}$. In such cases, $f_i$ simplifies to $f_i(x,\xi_i) = \frac{1}{n_i} \sum_{j=1}^{n_i} F_i(x,\xi_i^j)$. Our approach does not impose restrictive assumptions on the data distribution $\mathcal{D}_i$. In fact, we specifically focus on the extreme heterogeneous (non-i.i.d.) setting, where $\mathcal{D}_i \neq \mathcal{D}_{i^{'}}, \forall ~ i \neq i^{'}$ and the \textit{local optimal solution} $x_i^{\star}\leftarrow \argmin_{x \in \mathbb{R}^d} f_i(x)$ might significantly differ from the global minimizer of the objective function in \Eqref{eq: optimization-problem}. 

We are especially interested in the supervised classification task and let $\mathcal{M}_x: \mathcal{Z} \rightarrow \mathcal{Y}$ be a classifier parameterized by $x$. Here, $\mathcal{Z} \subseteq \mathbb{R}^D$ and $\mathcal{Y} = \{1, 2, \ldots, M\}$ denote the input and label spaces, respectively, $D$ is the input dimensionality, $M$ is the number of classes, and $d$ represents the number of parameters in the model $\mathcal{M}$. We set $F_i\left(x,\xi_i^j\right) = \mathcal{L}\left(\mathcal{M}_x \left(\mathbf{z}_i^j \right), y_i^j\right)$, where $\mathcal{L}$ is an appropriate loss function and $\xi_i^j \coloneq \left(\mathbf{z}_i^j, y_i^j\right)$ is a labeled training sample such that $\mathbf{z}_i^j \in \mathcal{Z}$ and $y_i^j \in \mathcal{Y}$. Furthermore, we mainly focus on the cross-silo FL setting ($N$ is small).

\paragraph{Federated Averaging (FedAvg).} A common approach for solving \Eqref{eq: optimization-problem} in the distributed setting is FedAvg \citep{mcmahan2017communication}. This algorithm involves the participants performing $K$ local steps of stochastic gradient descent (SGD) and communicating with the server over $T$ communication rounds. The server initializes the global model with $x_g^0$ and broadcasts it to all participants, which is then used to initialize the local models, i.e., $x_i^{1,0} = x_g^0$. In each communication round, the updates from the participants are averaged on the server and sent back to all participants. For a local step $k \in [K]$, communication round $t \in [T]$, and participant $i \in [N]$, the local and global iterates are updated as: 
\begin{equation}
    x_{i}^{t,k} = x_{i}^{t,k-1} - \eta_{\ell} \nabla f_i\left(x_i^{t,k-1}, \xi_i^{t,k-1}\right), \quad x_{i}^{t} = x_{i}^{t,K}, \text{ and  } x_g^{t} = x_g^{t-1} - \eta_g \left(x_g^{t-1} - \frac{1}{N} \sum_{i=1}^N x_{i}^{t}\right) , 
\end{equation}
\noindent where $\eta_{\ell}$ and $\eta_{g}$ are the local and global learning rates, respectively. The server then broadcasts the updated global model $x_g^{t}$ to all participants, which is then used to reinitialize the local models as $x_i^{t+1,0} = x_g^{t}$. 

In the FedAvg algorithm, the number of communication rounds necessary to achieve a certain precision is directly proportional to the heterogeneity measure \citep{li2019convergence}. Notably, this relationship holds true in convex settings; however, in non-convex scenarios, the algorithm may either not converge or may converge to a suboptimal solution. Stemming from this observation, our objective is to reduce the number of requisite communication rounds to achieve convergence, while simultaneously achieving a better solution. 

\begin{algorithm}[t]
    \caption{\texttt{FedPeWS} (For \colorbox{lightgreen}{\texttt{FedPeWS-Fixed}} variant, the steps highlighted in green are omitted and instead the server sets $m_i^{t} = m_i$, $\forall~t \in [W]$.)} 
    \renewcommand{\algorithmicrequire}{\textbf{Input:}}
    \renewcommand{\algorithmicensure}{\textbf{Return}} 
    \begin{algorithmic}[1] 
        \Require {Number of collaboration rounds $T$}, number of warmup rounds $W$, number of local steps $K$, local learning rate $\eta_{\ell}$, global learning rate $\eta_g$, mask learning rate $\eta_s$, $\lambda$ (mask diversity weight) 
        \State Initialize $x_g^0$ \colorbox{lightgreen}{and $s_g^0$, compute $\theta_g^0 = \sigma(s_g^0)$} 
        \For{$t = 1, \ldots, T$} 
            \If{$t > W$} // Use all parameters after warmup 
                \State Set $m_i^t = \mathbf{1}$, i.e., $m_i(\ell) = 1, ~ \forall ~ \ell \in [d]$ 
            \EndIf 
            \State Server sends global model $x_g^{t-1}$ \colorbox{lightgreen}{and global mask probability $\theta_g^{t-1}$} to all clients $i \in [N]$ 
            \For{client $i \in [N]$ in parallel}
                \State Initialize local model $x_{i}^{t,0} \gets x_g^{t-1}$  
                \State \colorbox{lightgreen}{$s_i^{t,0} \gets s_g^0$ \textbf{if } $t=1$ \textbf{else } $s_i^{t,0} \gets s_i^{t-1,K}$ \textbf{endif}} 
                \For{$k = 1, \ldots, K$} 
                    \State \colorbox{lightgreen}{\textbf{Procedure I}: Freeze model weights $x_{i}^{t,k-1}$}
                    \State \colorbox{lightgreen}{Optimize over $s$: $\mathcal{L}_s = f_i\Big(x_i^{t,k-1} \odot \mathcal{G}\left(s_i^{t,k-1}\right), \xi_i^{t,k-1}\Big) - \lambda \left\| \sigma\left(s_{i}^{t,k-1}\right) - {\theta_{g \backslash \{i\}}^{t-1}} \right\|_2^2$} 
                    \State \colorbox{lightgreen}{Update: $s_{i}^{t,k} \gets s_{i}^{t,k-1} - \eta_s \nabla_{s} \mathcal{L}_s$}
                    \State \textbf{Procedure II.} Freeze mask score vector $s_{i}^{t,k}$ 
                    \State Optimize over $x:$ $\mathcal{L}_x = f_i\Big(x_i^{t,k-1} \odot \text{\colorbox{lightgreen}{$\mathcal{G}\left(s_i^{t,k}\right)$}}, \xi_i^{t,k-1}\Big)$ 
                    \State Update: $x_{i}^{t,k} \gets x_{i}^{t,k-1} - \eta_{\ell} \nabla_{x} \mathcal{L}_x$ 
                \EndFor
                \State \colorbox{lightgreen}{Compute $m_i^{t} = \mathcal{G}(s_i^{t,K})$ and} upload $x_{i}^{t} \gets x_{i}^{t,K}$, \colorbox{lightgreen}{$m_i^{t}$} to server  
            \EndFor
            \State $x_g^{t} = x_g^{t-1} -  \eta_g \left(x_g^{t-1} - \dfrac{\sum_{i \in [N]}x_{i}^{t} \odot m_i^{t}}{\sum_{i \in [N]}m_i^{t}}\right)$ 
        \EndFor
    \end{algorithmic}
    \label{algorithm: adaptivesubnet}
\end{algorithm}

\section{Proposed FedPeWS Method}

\begin{figure}
    \centering
    \includegraphics[width=\linewidth]{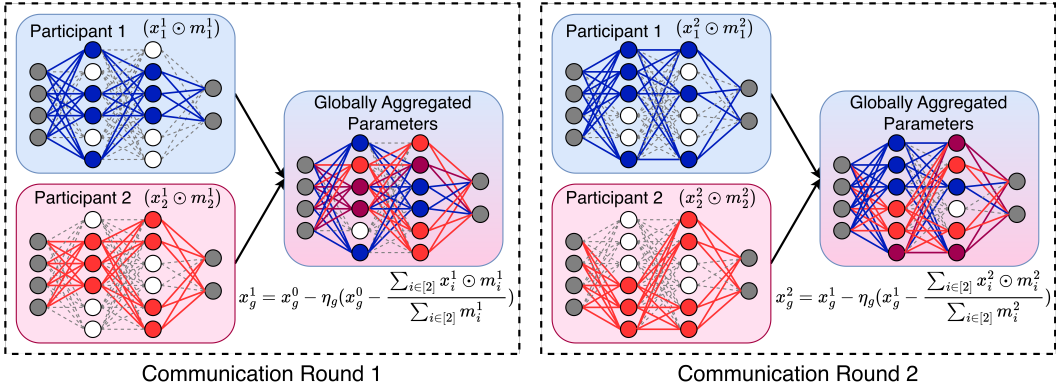} 
    \caption{Illustration of the proposed \texttt{FedPeWS} algorithm for two participants, which aggregates partial subnetworks ($x_i^t \odot m_i^t$) during the warmup phase to obtain a shared global model $x_g^t$. Here, $x_i^t$ and $m_i^t$ denote the local model and personalized mask of the $i^{\text{th}}$ participant in the $t^{\text{th}}$ round.}
    \label{fig: fedpews-masked-aggregation}
\end{figure}

The core idea of the proposed \texttt{FedPeWS} method is to allow participants to learn only a personalized subnetwork (a subset of parameters) instead of the entire network (all parameters) during the initial warmup phase. Let $m_i \in \{0,1\}^d$ be a binary mask vector denoting the set of parameters that are learned by participant $i$, $i \in [N]$. Note that $m_i(\ell) = 1$ indicates that the $\ell^{\textrm{th}}$ element of $x_i$ is selected for learning (value $0$ indicates non-selection), $\ell \in [d]$. Thus, during the warmup phase, the objective in \texttt{FedPeWS} is to learn the parameters $x$ along with the personalized masks $m_i$, i.e., 
\begin{equation}
    \label{eq: mask learning}
    \min_{x, \{m_i\}_{i\in[N]}} \frac{1}{N} \sum_{i=1}^N f_i(x \odot m_i),
\end{equation}
\noindent $\odot$ denotes element-wise multiplication. Note that $\mathcal{M}_{x \odot m_i}$ denotes the personalized subnetwork of participant $i$. When personalized masks are employed, the update rules can be modified as:
\begin{equation}
    x_{i}^{t,k} = x_{i}^{t,k-1} - \eta_{\ell} \nabla f_i\big(x_i^{t,k-1} \odot m_i^t, \xi_i^{t,k-1}\big) \; \text{and} \; x_g^{t} = x_g^{t-1} - \eta_g (x_g^{t-1} - \frac{\sum_{i \in [N]}x_{i}^{t} \odot m_i^t}{\sum_{i \in [N]}m_i^t}).
\end{equation}
The obvious questions regarding the \texttt{FedPeWS} method are: (i) how to learn these personalized masks $m_i$? and (ii) what should be the length of the warmup period?

\paragraph{Identification of personalized subnetworks:} It is not straightforward to directly optimize for the personalized binary (discrete) masks $m_i$ in \Eqref{eq: mask learning}. Hence, we make the following design choices. Firstly, personalized masks are learned at the neuron-level and then expanded to the parameter-level. Following IST \citep{yuan2022distributed}, masks are specifically applied only to the hidden layer neurons, while the head and tail neurons remain unaffected. However, unlike IST, the neuron-level masks are not randomly selected in each collaboration round. Instead, we learn real-valued personalized neuron-level mask score vectors $s_i \in \mathbb{R}^h$, which in turn can be used to generate the binary masks. Here, $h$ denotes the number of hidden neurons in the classifier $\mathcal{M}$ and $h \ll d$. A higher value of element $s_i(\ell)$, $\ell \in [h]$, indicates that the $\ell^{\textrm{th}}$ neuron is more likely to be selected by participant $i$.  Let $\mathcal{G}:\mathbb{R}^h \rightarrow \{0,1\}^d$ be the mask generation function that generates the binary parameter-level masks $m_i$ from neuron-level mask score vectors $s_i$, i.e., $m_i = \mathcal{G}(s_i)$. $\mathcal{G}$ consists of three steps. Firstly, we convert $s_i$ into probabilities by applying a sigmoid function, i.e., $\theta_i = \sigma(s_i)$, where $\theta_i \in [0,1]^h$ is the mask probability vector and $\sigma$ is the sigmoid function. Next, binary neuron masks $\tilde{m}_i$ are obtained by sampling from a Bernoulli distribution with parameter $\theta_i$, i.e., $\tilde{m}_i(\ell) \sim Bernoulli(\theta_i(\ell))$, $\forall~\ell \in [h]$. Finally, these binary neuron masks can be directly mapped to the binary parameter-level mask $m_i$, i.e., if a neuron is selected, all the weights associated with the selected neuron are also selected. Thus, \Eqref{eq: mask learning} can be reparameterized as:
\begin{equation}
    \label{eq: mask score learning}
    \min_{x, \{s_i\}_{i\in[N]}} \frac{1}{N} \sum_{i=1}^N f_i(x \odot \mathcal{G}(s_i)).
\end{equation}
The above equation can be optimized alternatively for the mask score vectors $s_i$ and the parameters $x$. The participants first optimize for the mask scores while the model parameters $x_{i}^{t,k}$ are frozen (Procedure I), and then switch to optimizing the model parameters while freezing the mask scores (Procedure II). In the mask training step (Procedure I), the optimization objective is defined as: 
\begin{equation}
    \mathcal{L}_s = f_i\Big(x_i^{t,k} \odot \mathcal{G}\left(s_i^{t,k}\right), \xi_i^{t,k-1}\Big) - \lambda \left\| \sigma\left(s_{i}^{t,k}\right) - \theta_{g \backslash \{i\}}^{t} \right\|_2^2; \qquad s_{i}^{t,k+1} \gets s_{i}^{t,k} - \eta_s \nabla_{s} \mathcal{L}_s, 
    \label{eq: mask-objective-func}
\end{equation}
where $\nabla_{s}$ indicates that the gradient is w.r.t. the mask score vector $s$, $\eta_s$ is the local learning rate for updating $s$, $\theta_g^t$ is the global mask probability at round $t$, $\theta_{g \backslash \{i\}}^t$ is the global mask probability excluding the probability mask of the current participant $i$, and $\lambda$ is the weight assigned to the mask diversity measure (second term). It is important to note that the personalized masks may not cover all neurons in the network. Maximizing the mask diversity measure encourages personalized masks to deviate as much as possible from the global mask, which facilitates better coverage of all the neurons in the global model. The diversity measure has an upper bound due to the sigmoid function: 
\begin{equation}
    \left\| \sigma\left(s_{i}^{t,k}\right) - \theta_{g \backslash \{i\}}^{t} \right\|_2^2 \leq h. 
\end{equation}

\noindent Given the difficulty in calculating $\nabla_s \mathcal{L}_s$ directly due to the discrete nature of Bernoulli sampling, we employ the straight-through estimator (STE) \citep{bengio2013STE} to approximate the gradients, which does not compute the gradient of the given function and passes on the incoming gradient as if the function were an identity function. 

During Procedure II, the optimization function for the model weights is expressed as: 
\begin{equation}
    \mathcal{L}_x = f_i\Big(x_i^{t,k} \odot \mathcal{G}\left(s_i^{t,k}\right), \xi_i^{t,k-1}\Big); \qquad x_{i}^{t,k+1} \gets x_{i}^{t,k} - \eta_{\ell} \nabla_{x} \mathcal{L}_x,
\end{equation}
where $\nabla_{x}$ indicates that the gradient is w.r.t. weights $x$. The \texttt{FedPeWS} algorithm alternates between these two procedures for $W$ rounds, where $W$ is the number of warmup rounds. 
At this point, the warmup stops and the participants switch to standard training for $(T-W)$ collaboration rounds. 
This approach ensures that each participant effectively contributes to the FL process while also tailoring the learning to their specific data distributions. The number of warmup rounds $W$ (or the proportion $\tau = \frac{W}{T}$) is a key hyperparameter of the \texttt{FedPeWS} algorithm, along with the weight $\lambda$ assigned to the mask diversity loss. While it would be ideal to have a principled method to select these hyperparameters, we use a grid search to tune them, which is currently a limitation. 

\paragraph{Use of fixed subnetworks:} When the number of participants is small and the data distributions of the participants are known apriori, the server can partition the full model into subnetworks of the same depth and assign a fixed subnetwork to each participant, i.e., $m_i^t = m_i$, $\forall~ t \in [W]$. Participants transmit only the masked updates back to the server during warmup, which then aggregates these masked parameters and redistributes them in their masked form. For the sake of utility, the server can design personalized masks such that the union of these masks covers all the neurons. This variant of \texttt{FedPeWS} is referred to as \texttt{FedPeWS-Fixed} and follows the same algorithm in \Algref{algorithm: adaptivesubnet}, except for the omission of the highlighted (in green color) steps.

\section{Experiments and Results}
\label{section: exp-and-results}

\subsection{Datasets and Network Architecture} 
\label{section: datasets}

\begin{wrapfigure}{r}{0.52\textwidth}
    \vspace{-1em}
    \centering
    \includegraphics[width=\linewidth]{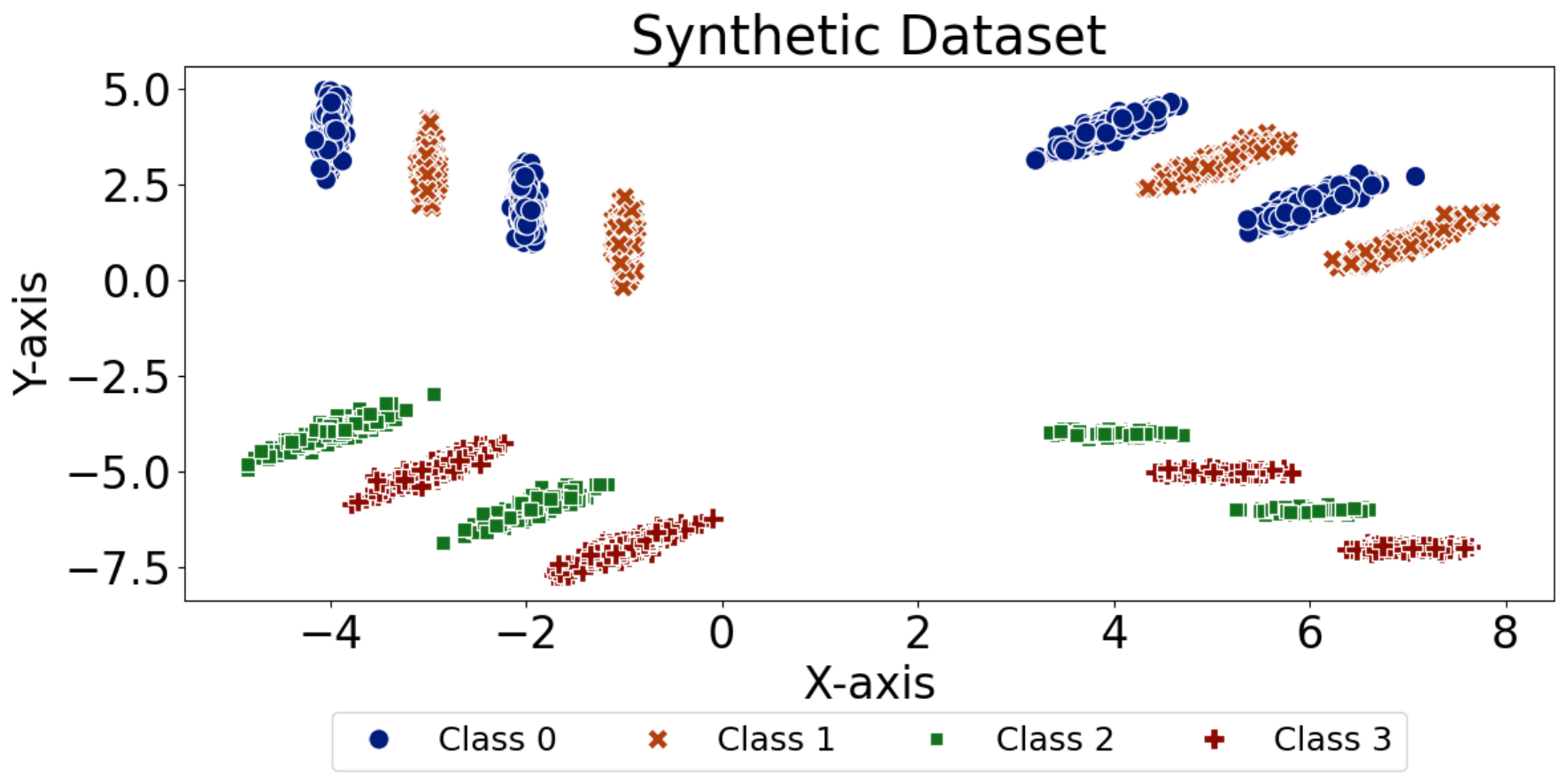} 
    \caption{Samples from the custom synthetic dataset.}
    \label{fig:synthetic-data}
\end{wrapfigure}
\paragraph{Synthetic Dataset.} To effectively evaluate the performance of the proposed algorithm, we generated a custom synthetic dataset that simulates the extreme non i.i.d. scenario. This dataset encompasses four classes, each characterized by four 2D clusters determined by specific centers and covariance matrices. Note that the clusters from different classes interleave each other as shown in \Figref{fig:synthetic-data}.  For this dataset, we utilize a neural network consisting of five fully-connected (FC) layers, each followed by ReLU activation functions, except the last layer. To enhance the dataset complexity and aid FC network learning, we transform these 2D points into 5D space using the transformation $[x, y, x^2, y^2, xy]$, based on their $(x,y)$ coordinates. We generate two versions of this dataset, \textbf{Synthetic-32K} and \textbf{Synthetic-3.2K}, depending on the number of data points in the training set. The former has $32000$ samples, with each class containing $8000$ data points, while the latter has ten times fewer data points. 

\paragraph{CIFAR-MNIST.} We integrate two distinct datasets, CIFAR-10 \citep{krizhevsky2009learning} and MNIST \citep{lecun1998mnist}, to explore how different clients might adapt when faced with disparate data sources. CIFAR-10 comprises $32 \times 32$ pixel images categorized into 10 object classes. MNIST, typically featuring $28 \times 28$ pixel images across 10 digit classes, is upscaled to $32 \times 32$ pixel to standardize dimensions with CIFAR-10. 
We compile a balanced dataset by randomly selecting 400 samples from each class for the training set and 200 samples for the test set from the combined pool of 20 classes. This setup aims to simulate a FL environment where multiple clients handle significantly varied data types. For this dataset, we employ a convolutional neural network comprising four convolutional layers, each having a kernel size of 3 and padding of 1, followed by max pooling. This is succeeded by a fully connected layer. This architecture was used because of its simplicity and widespread use in the literature \citep{yuan2022distributed, isik2022sparse}. 

\paragraph{\{Path-OCT-Tissue\}MNIST.} We amalgamate three distinct medical datasets: PathMNIST, OCTMNIST, and TissueMNIST \citep{medmnistv2}, to develop a universal medical prognosis model capable of recognizing various tasks using a single model. The datasets contain 9, 4, and 8 classes, respectively, totaling 21 classes. For this dataset, we utilized the same architecture and training details described in the CIFAR-MNIST dataset.

\subsection{Experimental Setup} 
\label{subsection: exp-setup}

\paragraph{Dataset partitioning.} For scenarios with a smaller number of collaborators ($N={2, 3, 4}$), we manually partition the training dataset to tailor the data distribution to specific participants. In the $N=2$ scenario, we partition as follows: (i) For the Synthetic dataset, encompassing both Synthetic-32K and Synthetic-3.2K, even-numbered classes are assigned to participant 1, while odd-numbered classes are allocated to participant 2. (ii) For the CIFAR-MNIST combination, all CIFAR-10 samples are assigned to participant 1, with MNIST samples allocated to participant 2. In the $N=3$ scenario, the \{Path-OCT-Tissue\}MNIST dataset is partitioned into three splits corresponding to the individual datasets, with PathMNIST assigned to participant 1, OCTMNIST to participant 2, and TissueMNIST to participant 3. For the $N=4$ scenario, the synthetic dataset is divided so that each class is exclusively allocated to one of the four participants. 
For scenarios with a larger number of participants ($N \geq 10$), we employ a Dirichlet distribution to partition the training set. This approach utilizes a concentration parameter $\alpha$ to simulate both homogeneous and heterogeneous data distributions \citep{yurochkin2019bayesian, li2020practical, lin2020ensemble, wang2020federated}. We experiment with various values of $\alpha$, specifically $\alpha \in \{0.1, 0.5, 1.0, 2.0, 5.0\}$, to explore the effects of dataset heterogeneity (lower $\alpha$ values) and homogeneity (higher $\alpha$ values) on the model performance. This methodological diversity allows us to comprehensively assess our approach under varying data conditions. Results for large $N$ ($> 100$) are reported in the appendix.  

\paragraph{Training details.} 
In federated optimization, we primarily benchmark against the FedAvg algorithm \citep{mcmahan2017communication}, a standard approach in federated learning. However, our algorithm is designed to be versatile, functioning as a `plug-and-play' solution that is compatible with various other optimizers. To demonstrate this adaptability, we also conduct experiments using FedProx \citep{li2020federated}, showcasing our method's capabilities across different optimization frameworks. Moreover, we compare these two variants to SCAFFOLD \citep{karimireddy2020scaffold}, FedNova \citep{wang2020tackling}, and MOON \citep{li2021moon}. For our experiments, we fix the local learning rate $\eta_{\ell}=0.001$ in the Synthetic-32K dataset case, and we set $\eta_{\ell}=0.01$ for other experiments. Also, the mask learning rate is fixed $\eta_s=0.1$. Furthermore, we vary the global learning rate $\eta_g \in \{0.1, 0.25, 0.5, 1.0\}$ to observe the differences in optimization behavior between the baseline and our proposed methods. Additionally, we employ two distinct batch sizes $\{32, 8\}$ for Synthetic-32K and Synthetic-3.2K, respectively. For experiments involving the CIFAR-MNIST and \{Path-OCT-Tissue\}MNIST datasets, we standardize the batch size to 64. 
We conduct our experiments on NVIDIA RTX A6000 GPUs on an internal cluster server, with each run utilizing a single GPU. The execution time for each run is capped at less than an hour, which indicates the maximum execution time rather than the average. All results are averaged over three independent runs and the average accuracy is reported on the global test dataset.

\begin{table}[t]
    \centering
    \caption{The required number of collaboration rounds to reach target accuracy $\upsilon$ $\%$ and the final accuracy after $T$ rounds. The results are averaged over 3 seeds. 
    {\color{red} $\times$} indicates that the algorithm cannot reach target accuracy $\upsilon$ within $T$ rounds and {\color{red} NA} means that it reaches $\upsilon$ only in one random seed. } 
    \resizebox{\linewidth}{!}{ 
        \begin{tabular}{cr|ccccc} 
        \toprule
        \rowcolor{lightgray} \multicolumn{2}{r}{\textbf{Dataset $/$ Batch size}} & \multicolumn{3}{c}{Synthetic-32K, $32$} & \multicolumn{1}{c}{Synthetic-3.2K, $8$} 
        \\ \midrule 
        \rowcolor{lightgray} \multicolumn{2}{r}{\textbf{Parameters} $\{\eta_g / \lambda / \tau\}$} & $\{1.0 / 5.0 / 0.125\}$ & $\{0.5/2.0/0.2\}$ & $\{0.25/1.0/0.1875\}$ & \multicolumn{1}{c}{$\{0.1/2.0/0.1\}$} \\ \midrule
        \rowcolor{lightgray} \multicolumn{2}{r}{\textbf{Target accuracy $\upsilon (\%)$}} & $99$ & $90$ & $75$ & $99$ 
        \\ \midrule 
        \multirow{2}{*}{\parbox{3.5cm}{\centering \textbf{No. of rounds to \\ reach target accuracy}}} & FedAvg & $148\pm3.79$ & \color{red} \cellcolor{lightred} $199\pm \text{NA}$ & \color{red} \cellcolor{lightred} $\times$ & \color{red} \cellcolor{lightred} $371\pm\text{NA}$ 
        \\ \cmidrule{2-6}
         & \cellcolor{lightblue} \texttt{FedAvg+PeWS} & \cellcolor{lightblue} $\mathbf{115\pm7.21}$ & \cellcolor{lightblue} $\mathbf{182\pm6.81}$ & \cellcolor{lightblue} $\mathbf{286\pm7.93}$ & \cellcolor{lightblue} $\mathbf{301\pm10.59}$ 
        \\ \midrule 
        \multirow{2}{*}{\parbox{3.75cm}{\centering \textbf{Final accuracy after \\ $T$ collaboration rounds}}} & FedAvg & \multicolumn{1}{c}{$99.94\pm0.05$} & $91.40\pm7.25$ & $67.64\pm0.90$ & $97.33\pm3.89$ 
        \\ \cmidrule{2-6}
        & \cellcolor{lightblue} \texttt{FedAvg+PeWS} & \cellcolor{lightblue} $\mathbf{99.96\pm0.01}$ & \cellcolor{lightblue} $\mathbf{99.49\pm0.60}$ & \cellcolor{lightblue} $\mathbf{83.50\pm3.52}$ & \cellcolor{lightblue} $\mathbf{99.66\pm0.19}$ 
        \\ \bottomrule 
        \end{tabular}
    }
    \label{table: convergence-table}
\end{table}

\begin{figure}[t] 
    \centering
    \includegraphics[width=\linewidth]{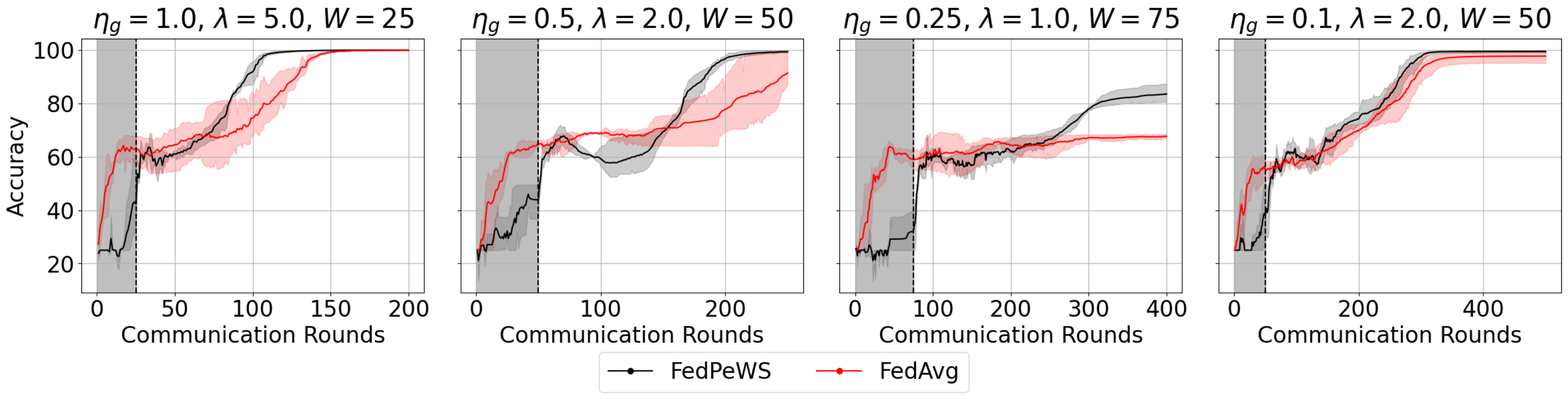}
    \caption{Results on Synthetic-\{32, 3.2\}K datasets with batch sizes \{32, 8\}, global learning rates $\eta_g \in \{1.0, 0.5, 0.25, 0.1\}$ and communication rounds $T \in \{200, 250, 400, 500\}$. See Table \ref{table: convergence-table} for details. \texttt{FedPeWS} consistently converges faster and outperforms FedAvg.} 
    \label{fig: synthetic-all}
\end{figure}

\subsection{Experimental Results}
\label{subsection: exp-results}

\begin{figure}[t]
    \centering
    \begin{subfigure}[t]{0.495\linewidth}
        \centering
        \includegraphics[width=\linewidth]{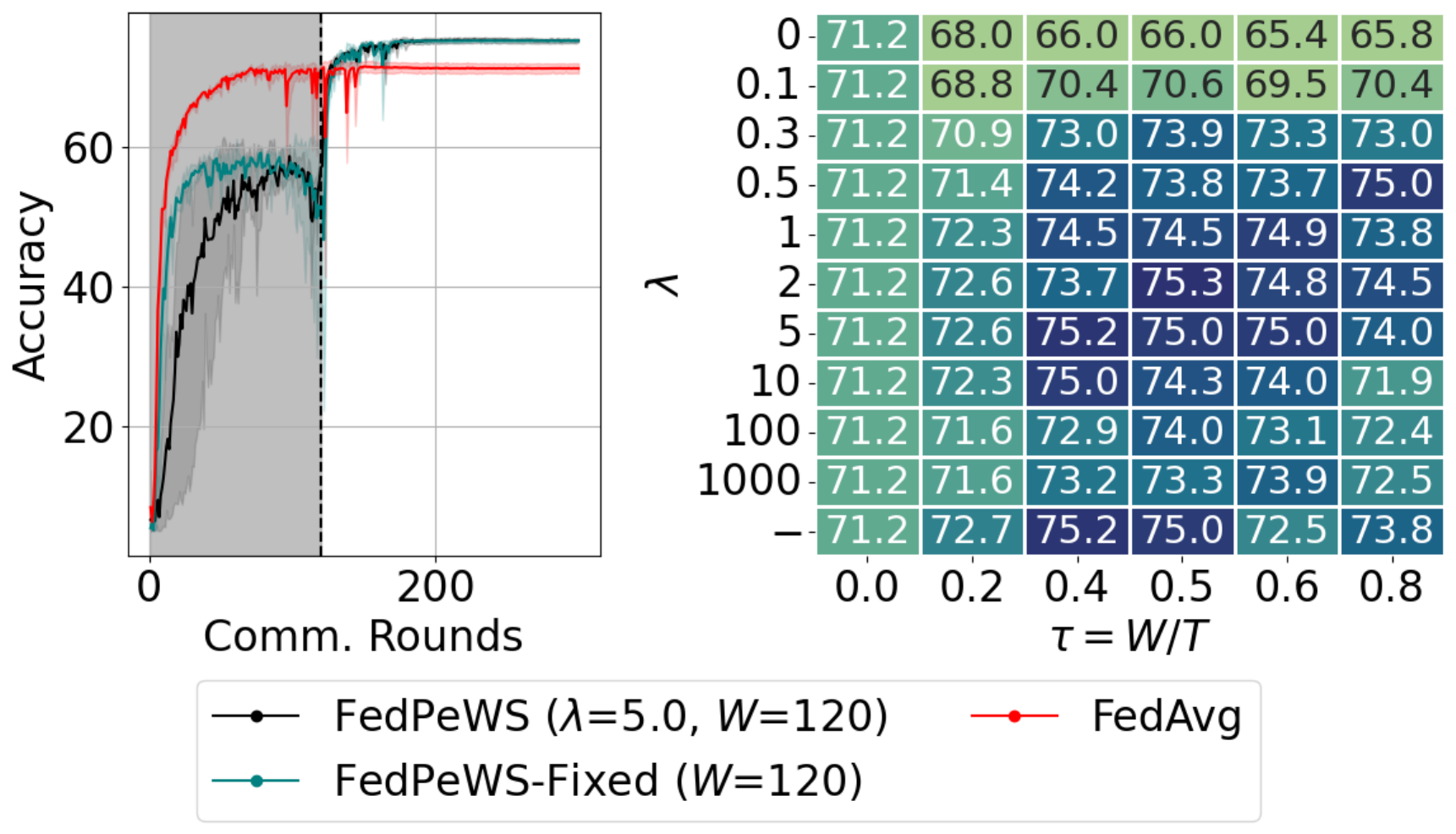}
        \caption{CIFAR-MNIST dataset.}
        \label{fig: cifar-mnist-figure-heatmap}
    \end{subfigure}
    \hfill
    \begin{subfigure}[t]{0.495\linewidth}
        \centering
        \includegraphics[width=\linewidth]{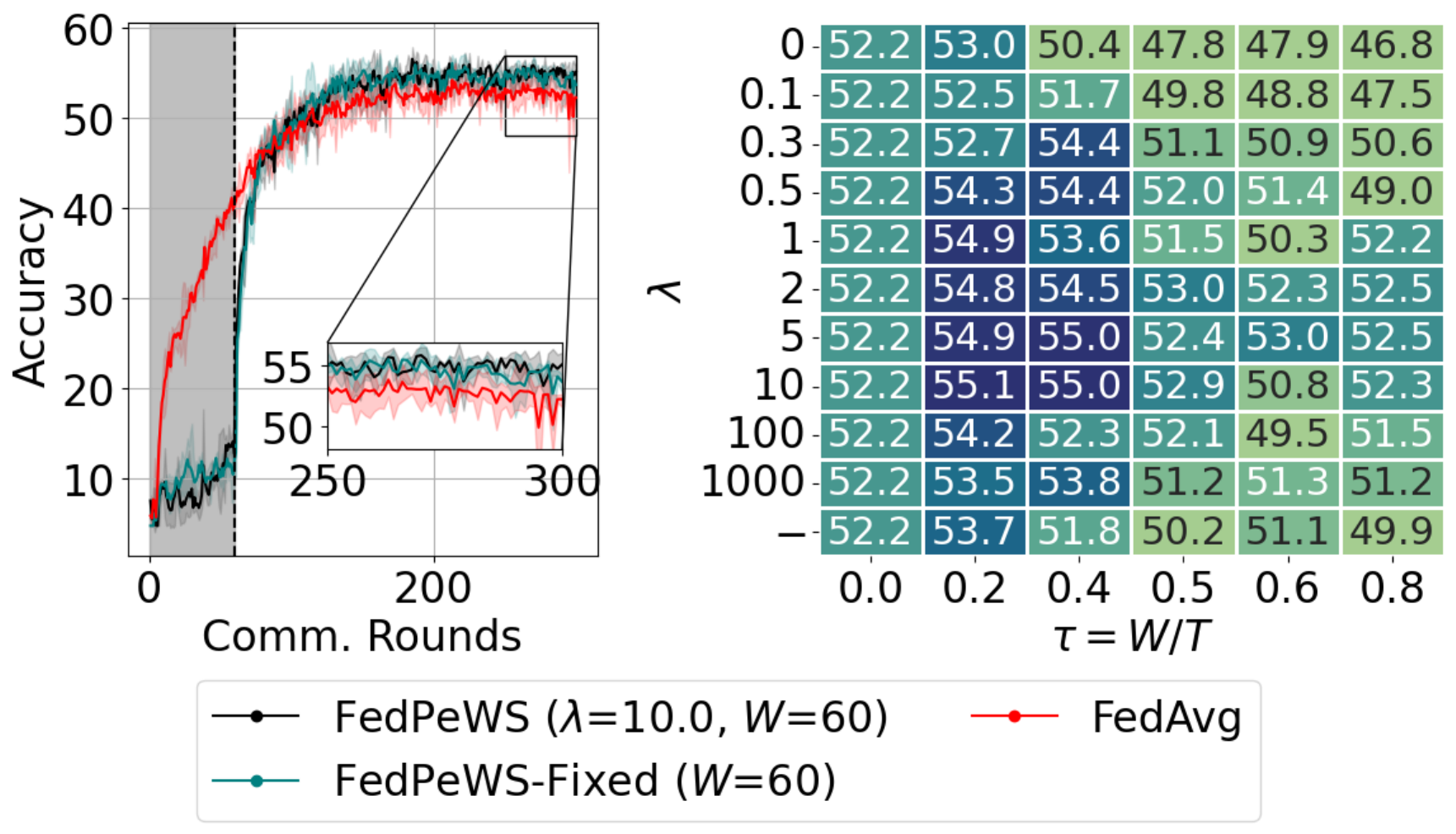}
        \caption{\{Path-OCT-Tissue\}MNIST dataset.}
        \label{fig: pot-figure-heatmap}
    \end{subfigure}
    \caption{Results for experiments with $T=300$ on (a) CIFAR-MNIST and (b) \{Path-OCT-Tissue\}MNIST datasets. \textbf{(a)} Participant 1 uses MNIST; Participant 2 uses CIFAR-10; ablation study for $\lambda$ and $\tau$ (Table \ref{tab: cifar-mnist-ablation}). \textbf{(b)} $N=3$ participants use \{PathMNIST, OCTMNIST, TissueMNIST\}; ablation study for $\lambda$ and $\tau$ (Table \ref{tab: pot-ablation}). \texttt{FedPeWS-Fixed} results appear in the last row; $\tau=0.0$ denotes FedAvg.} 
    \label{fig: cifar-mnist-pot-accuracy-heatmap} 
\end{figure}

Our experimental analysis focuses on assessing the performance of our proposed \texttt{FedPeWS} algorithm within the FL framework. The key findings from our studies are as follows: (i) The \texttt{FedPeWS} approach demonstrates a significant reduction in the number of communication rounds required to achieve target accuracy while also enhancing the final accuracy post-convergence. (ii) The \texttt{FedPeWS} algorithm is robust across different levels of data heterogeneity. (iii) In scenarios where full knowledge of the participant data distributions is available, the server can employ \texttt{FedPeWS-Fixed} method (Figure \ref{fig: cifar-mnist-pot-accuracy-heatmap}). 

\paragraph{Improved communication efficiency and accuracy.} 

\begin{wrapfigure}{r}{0.51\textwidth}
    \vspace{-1.2em} 
    \centering
    \includegraphics[width=\linewidth]{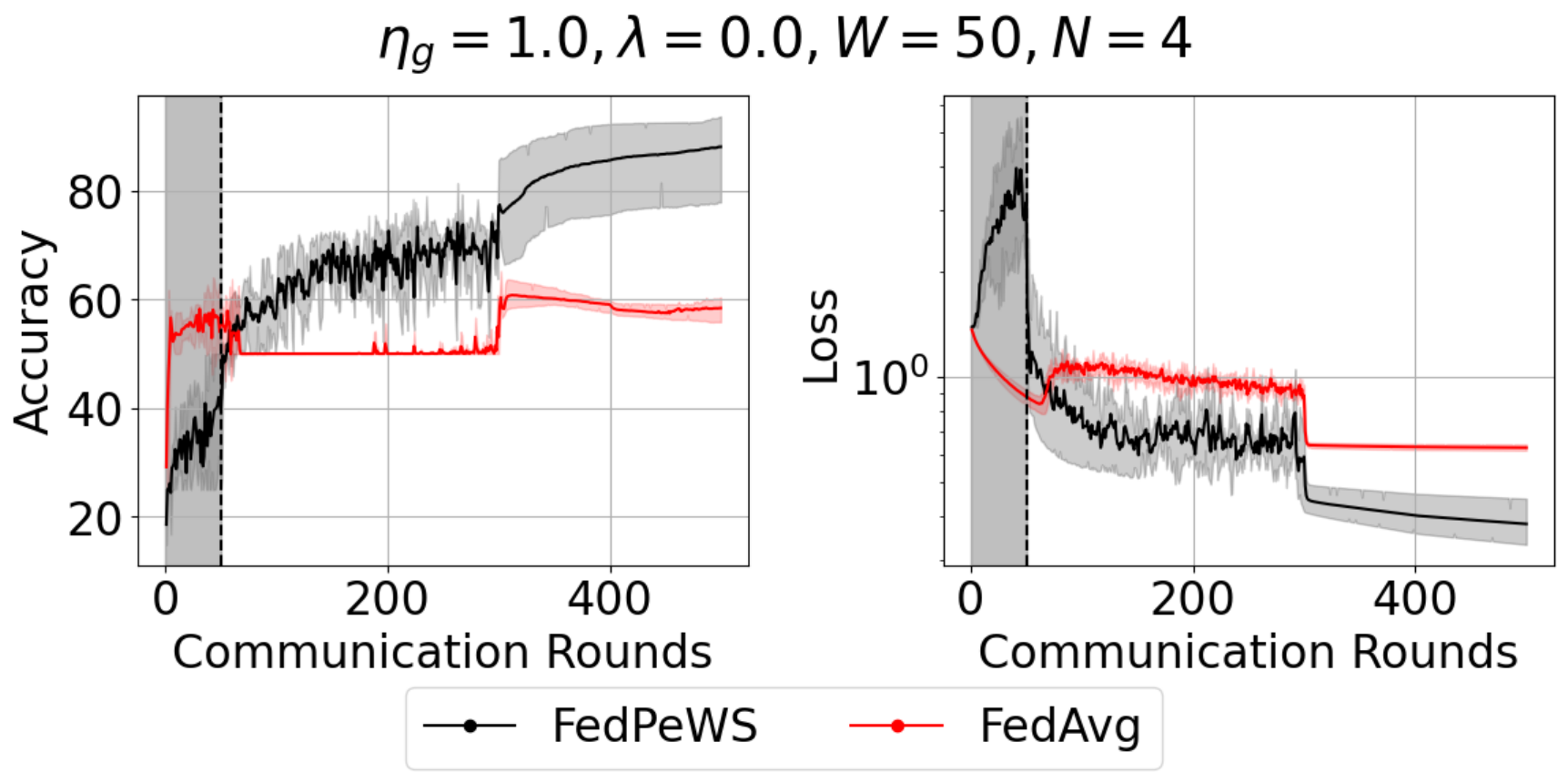}
    \caption{Visualization of validation accuracy and loss on the Synthetic-32K dataset with $N=4$. 
    } 
    \label{fig: n4-synthetic}
\end{wrapfigure} 

We initially report the required number of communication rounds to reach the target accuracy and the final accuracy after $T$ communication rounds for the synthetic dataset in \Tabref{table: convergence-table}. The results underscore that the incorporation of a personalized warmup phase in a federated setup significantly reduces the required number of communication rounds across all tested scenarios. Notably, in specific instances, such as with the Synthetic-32K dataset and $\eta_g=0.25$, the conventional FedAvg algorithm does not meet the target accuracy within the $T$ communication rounds. Conversely, in scenarios where $\eta_g \in \{0.25, 0.1\}$, FedAvg only achieves the target accuracy in one of the seeds, exhibiting suboptimal performance in the other two runs. From \Figref{fig: synthetic-all}, it is evident that our proposed \texttt{FedPeWS} algorithm surpasses FedAvg in both communication efficiency and accuracy. 
We also consider a more extreme data heterogeneity scenario with $N=4$ participants, depicted in Figure \ref{fig: n4-synthetic}, where FedAvg completely fails by reaching only $58.4\pm2.33\%$, whereas our \texttt{FedPeWS} approach reaches $91.13\pm3.55\%$ accuracy by significantly outperforming the base optimizer (FedAvg) with a gain of $\mathbf{32.72\%}$. It is crucial to highlight that in this experiment, we set $\lambda=0.0$, effectively not enforcing diversity as outlined in Equation \ref{eq: mask-objective-func}. This approach focuses solely on optimizing the masks using the first loss component, which depends only on the data distributions of each participant. This shows that, in specific scenarios, we can learn the personalized masks (Procedure I) without the need to adjust the $\lambda$ parameter, while still achieving a better performance than the base optimizer.

\paragraph{Sensitivity to $\lambda$ and $\tau$ parameters.} \Figref{fig: cifar-mnist-pot-accuracy-heatmap} showcases the results of experiments on the CIFAR-MNIST dataset with $N=2$ participants and \{Path-OCT-Tissue\}MNIST dataset with $N=3$ participants. 
The left-side plots of Figures \ref{fig: cifar-mnist-figure-heatmap} and \ref{fig: pot-figure-heatmap}, which depict the performance of the global model (averaged over 3 runs), demonstrate that our method consistently achieves higher accuracy. The right side figures feature heatmap plots that annotate the global model accuracy obtained varying $\lambda \in [0, 1000]$ and $\tau \in [0.0, 0.8]$ parameters. An additional row $(\lambda=-)$ represents the \texttt{FedPeWS-Fixed} approach, where user(server)-defined fixed masks are employed. In this method, we simply split the full network into $N$ partitions, with each partition assigned to a participant (for detailed instructions see Section \ref{subsection: fixed-mask-how-to} in the appendix). The results indicate that our approach has a low sensitivity to variations in $\lambda$ and $\tau$. For more detailed insights, please refer to Tables \ref{tab: cifar-mnist-ablation} and \ref{tab: pot-ablation} in the appendix.

\begin{wraptable}{r}{0.52\textwidth} 
    \vspace{-1.25em} 
    \centering 
    \caption{Comparison to the SOTA algorithms.} 
    \resizebox{\linewidth}{!}{ 
        \begin{tabular}{r|cc} 
            \toprule 
            \rowcolor{lightgray} \textbf{Dataset} & \textbf{CIFAR-MNIST} & \textbf{\{P-O-T\}MNIST} \\ \midrule 
            Fedavg                       & $71.78 \pm 0.66$ & $52.83 \pm 1.26$ \\ 
            FedProx                      & $72.27 \pm 0.88$ & $51.28 \pm 1.03$ \\ 
            SCAFFOLD                     & $71.83 \pm 0.24$ & $53.05 \pm 0.60$ \\ 
            FedNova                      & $71.63 \pm 0.98$ & $53.05 \pm 0.83$ \\ 
            MOON                         & $71.84 \pm 1.09$ & $52.10 \pm 0.19$ \\ \midrule 
            \rowcolor{lightblue} \texttt{FedAvg+PeWS} & $\mathbf{75.83 \pm 0.88}$ & $\mathbf{55.12 \pm 0.56}$ \\ 
            \rowcolor{lightblue} \texttt{FedProx+PeWS} & $75.04 \pm 0.85$ & $54.67 \pm 0.43$ \\ \bottomrule 
        \end{tabular}
    }
    \label{tab: comparison-to-sota}
\end{wraptable}

\paragraph{Comparison to the state-of-the-art (SOTA) algorithms.} We evaluate our proposed algorithm, FedPeWS, against state-of-the-art (SOTA) algorithms, with the results summarized in Table \ref{tab: comparison-to-sota}. As highlighted in the table, algorithms such as FedProx, SCAFFOLD, FedNova, and MOON demonstrate performance comparable to FedAvg, as illustrated in Figure \ref{fig: cifar-mnist-pot-accuracy-heatmap}. 
In contrast, our algorithm, when applied on top of FedAvg and FedProx, achieves superior performance.

\section{Conclusion}
In this work, we introduced a novel concept called \textit{personalized warmup via subnetworks} for heterogeneous FL -- a strategy that enhances convergence speed and can seamlessly integrate with existing optimization techniques. Results demonstrate that the proposed \texttt{FedPeWS} approach achieves higher accuracy than the relevant baselines, especially when there is extreme statistical heterogeneity.

\section*{Acknowledgements}

This material is partly based on work supported by the Office of Naval Research N00014-24-1-2168.



\bibliography{reference}


\newpage 
\appendix 

\tableofcontents

\newpage 
\section{Scalability of the Mask Mechanism}
Our approach is designed as a plug-and-play extension to existing federated optimization methods with minimal overhead. 

To estimate the computational overhead, let us assume that a baseline method such as FedAvg incurs a training cost of 1 unit per communication round. In our method, a warm-up phase occupies a fraction $\tau$ of the total communication rounds (e.g., $\tau \in \{0.1, 0.5\}$). Let $\omega$ be the computational complexity of each warmup round. Consequently, the total cost of the overall training will increase only when $\omega$ is greater than 1. During the warmup rounds, additional costs from mask learning are incurred. It must be noted that we learn only neuron-level masks, and the number of hidden neurons is much lower than the number of parameters. On the other hand, these additional mask learning costs are offset by the fact that only sparse subnetworks are learned during the warmup rounds. Empirically, the observed sparsification level lies in the range $[0.1, 0.5]$. Under the worst-case scenario of sparsification at 0.5, these two factors (computational cost increase due to mask learning and computational cost decrease due to sparsification) cancel each other out and $\omega \approx 1$. Therefore, the computational cost of \texttt{FedPeWS} is approximately the same as FedAvg when sparsification is $0.5$. For more aggressive sparsification ($< 0.5$), \texttt{FedPeWS} becomes more computationally efficient than FedAvg. 

\section{Additional Experimental Details} 
\label{suppl-section: arch}

\subsection{Network Architecture Details} 
The network for the synthetic dataset (detailed in Section \ref{subsection: exp-setup}) consists of five fully connected (FC) layers, each followed by  ReLU activation functions, except for the last layer. We provide the details of this architecture in Table \ref{tab: synthetic-arch}. 

\begin{table}[th]
    \centering
    \caption{Architecture for synthetic dataset models used in the experiments.}
    \resizebox{0.75\linewidth}{!}{ 
        \begin{tabular}{lllllll} 
            \toprule 
            \cellcolor{lightgray}\textbf{Layer} & Input & FC1 & FC2 & FC3 & FC4 & FC5 \\ \midrule 
            \cellcolor{lightgray}\textbf{Dimensions} & [5] & [5, 32] & [32, 64] & [64, 128] & [128, 32] & [32, 4] \\ \bottomrule
        \end{tabular}
    }
    \label{tab: synthetic-arch}
\end{table}

The network for the CIFAR-MNIST and \{Path-OCT-Tissue\}MNIST datasets includes three convolutional layers followed by max pooling, and a fully connected layer. The details of this architecture is provided in Table \ref{tab: conv-arch}. 
\begin{table}[H]
    \centering
    \caption{Architecture for CIFAR-MNIST dataset models. Every convolutional layer is followed by a max pooling layer with kernel size 2 and stride 2. }
    \resizebox{\linewidth}{!}{ 
        \begin{tabular}{lllllll}
            \toprule
             \cellcolor{lightgray}\textbf{Layer} & Input & Conv1 & Conv2 & Conv3 & Flatten & FC \\ \midrule 
             \cellcolor{lightgray}\textbf{Dimensions} & [3, 32, 32] & [3, 32, 3, 3] & [32, 64, 3, 3] & [64, 128, 3, 3] & [2048] & [2048, 20] \\ \bottomrule 
        \end{tabular}
    }
    \label{tab: conv-arch}
\end{table}

\subsection{Fixed Mask Generation} 
\label{subsection: fixed-mask-how-to}
\Figref{fig: fixed-mask-split} illustrates how we design masks for \texttt{FedPeWS-Fixed} experiments in scenarios with $N=2$ participants. For cases involving $N=4$ participants, the full network $\mathcal{M}_x$ (classifier $\mathcal{M}$ parameterized with $x$) is divided into four subnetworks, vertically, with each subnetwork corresponding to one of the participants. As such, we vertically partition the hidden neurons in the network into $N$ groups (subnetworks) and design the mask to assign each group to one participant, ensuring no overlap. This design choice is based on the assumption that classes held by each participant are highly heterogeneous, thus preventing any intersection in the masks. This setting is specifically tailored for the \texttt{FedPeWS-Fixed} method and doesn't necessitate performing optimization over the masks $m_i$; they are kept fixed. 

\begin{figure}[h] 
    \centering
    \includegraphics[width=0.9\textwidth]{images/fixedmask_V1-500.pdf}
    \caption{Illustration of manual mask setting in the \texttt{FedPeWS-Fixed} method. The left figure illustrates the complete network with all neurons active and full connections. The middle figure represents subnetwork 1, utilizing only the left portion of the full network, where $m_1$ corresponds to this left side. Conversely, the right figure indicates the part of the network used for subnetwork 2. This setting is employed in all experiments involving $N=2$ participants.} 
    \label{fig: fixed-mask-split}
\end{figure}

\section{Experimental Results}

\subsection{Varying Degrees of Heterogeneity} 
\begin{figure}[h] 
    \centering
    {\includegraphics[trim={0cm 0.2cm 0 0.2cm},clip,width=\linewidth]{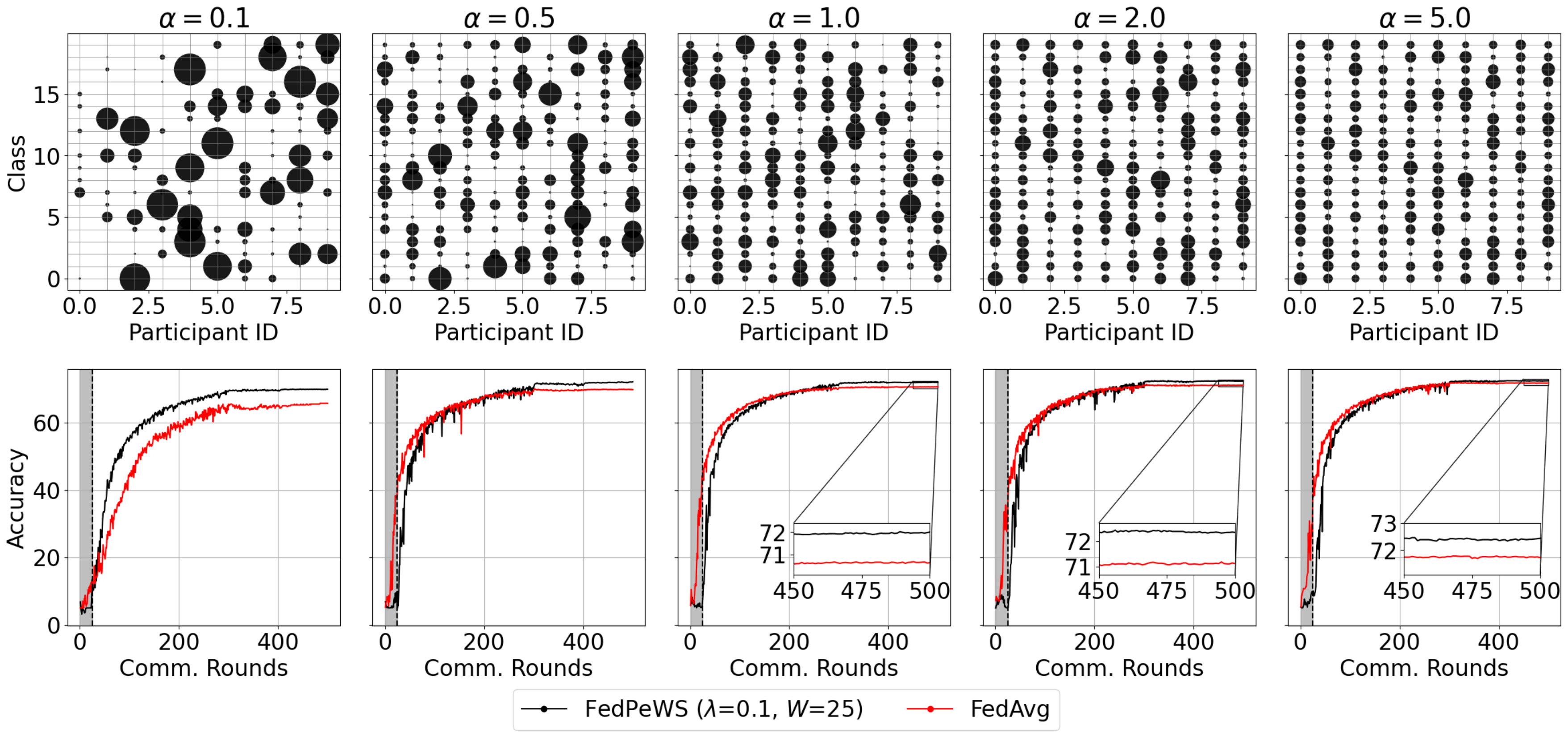}}
    \caption{Top: illustration of number of samples per class allocated to each client, that is indicated by dot sizes, for different concentration $\alpha$ values. Bottom: visualization of the experiments on CIFAR-MNIST dataset with $N=10$ participants with different levels of heterogeneity.} 
    \label{fig: dirichlet}
\end{figure}

\Figref{fig: dirichlet} demonstrates that our \texttt{FedPeWS} approach consistently outperforms FedAvg, with gains directly related to the degree of data heterogeneity. The figure clearly shows that the advantage of using our method is more pronounced under conditions of high data heterogeneity. As heterogeneity levels decrease, our method becomes comparable to FedAvg. 

\subsection{Comparison to FedProx} 
\begin{figure}[t]
    \centering
    \includegraphics[trim={0cm 0.2cm 0 0.2cm},clip,width=\linewidth]{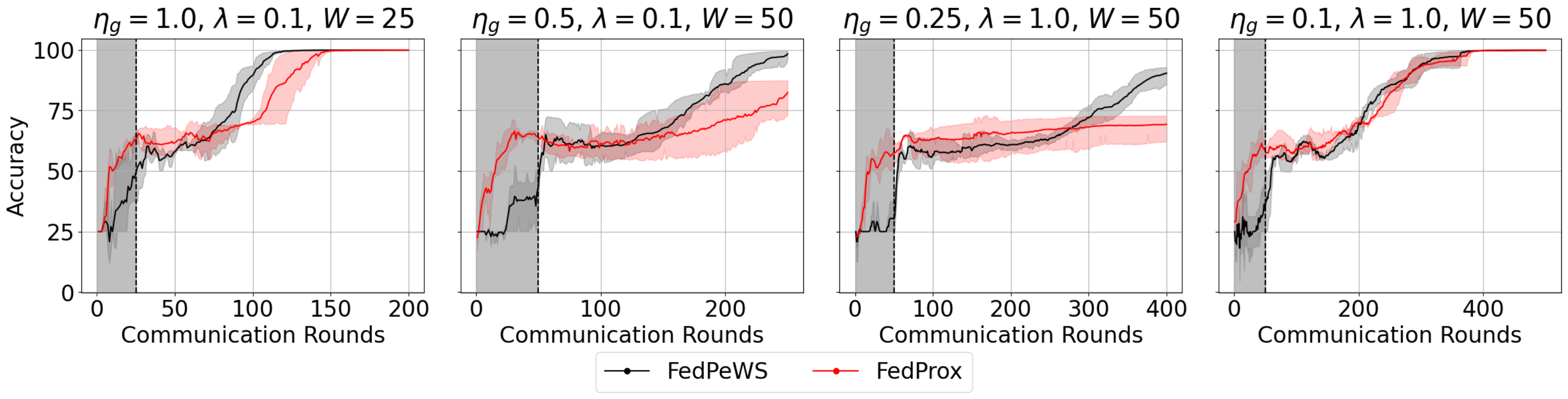} 
    \caption{Comparison of our proposed method and FedProx \citep{li2020federated} on Synthetic-\{32, 3.2\}K datasets. Refer to Table \ref{table: convergence-table-fedprox} for the corresponding numbers.} 
    \label{fig:fedprox-scenario13}
\end{figure}

We also present results using the FedProx optimizer on Synthetic-32K and Synthetic-3.2K datasets in Figure \ref{fig:fedprox-scenario13} and Table \ref{table: convergence-table-fedprox}, employing global learning rates $\eta_g = \{1.0, 0.5, 0.25, 0.1\}$. Note that we adapt Algorithm \ref{algorithm: adaptivesubnet} to incorporate the FedProx algorithm as the base optimizer, instead of FedAvg. 
We selected the best performing proximal term scaler $0.01$ after tuning and evaluating different values from a set of potential values $\{0.001, 0.01, 0.1, 0.5\}$, based on the findings in \citep{li2020federated}. The results demonstrate that \texttt{FedPeWS} outperforms FedProx in terms of both communication efficiency and final accuracy across the tested scenarios, except the last scenario (Synthetic-3.2K dataset with batch size 8 and $\eta_g=0.1$), where the performance of \texttt{FedPeWS} is comparable to that of FedProx. 

\begin{table}[t]
    \centering
    \caption{The required number of collaboration rounds to reach target accuracy $\upsilon$ $\%$ using FedProx algorithm and the final accuracy after $T$ rounds. The results are averaged over 3 seeds. 
    {\color{red} $\times$} indicates that the algorithm cannot reach target accuracy $\upsilon$ within $T$ rounds.} 
    \resizebox{\linewidth}{!}{ 
        \begin{tabular}{cr|ccccc} 
        \toprule
        \rowcolor{lightgray} \multicolumn{2}{r}{\textbf{Dataset $/$ Batch size}} & \multicolumn{3}{c}{Synthetic-32K, $32$} & \multicolumn{1}{c}{Synthetic-3.2K, $8$} 
        \\ \midrule 
        \rowcolor{lightgray} \multicolumn{2}{r}{\textbf{Parameters} $\{\eta_g / \lambda / \tau\}$} & $\{1.0 / 0.1 / 0.125\}$ & $\{0.5/0.1/0.2\}$ & $\{0.25/1.0/0.1875\}$ & \multicolumn{1}{c}{$\{0.1/1.0/0.1\}$} \\ \midrule
        \rowcolor{lightgray} \multicolumn{2}{r}{\textbf{Target accuracy $\upsilon (\%)$}} & $99$ & $90$ & $75$ & $99$ 
        \\ \midrule 
        \multirow{2}{*}{\parbox{3.5cm}{\centering \textbf{No. of rounds to \\ reach target accuracy}}} & FedProx & $138\pm13.22$ & \cellcolor{lightred} \color{red} $\times$ & \cellcolor{lightred} \color{red} $\times$ & $362\pm20.00$ 
        \\ \cmidrule{2-6} 
         & \cellcolor{lightblue} \texttt{FedProx+PeWS} & \cellcolor{lightblue} $\mathbf{115\pm5.29}$ & \cellcolor{lightblue} $\mathbf{211\pm16.52}$ & \cellcolor{lightblue} $\mathbf{314\pm27.83}$ & \cellcolor{lightblue} $\mathbf{344\pm27.30}$ 
        \\ \midrule 
        \multirow{2}{*}{\parbox{3.75cm}{\centering \textbf{Final accuracy after \\ $T$ collaboration rounds}}} & FedProx & \multicolumn{1}{c}{$99.95\pm0.02$} & $82.43\pm7.98$ & $69.26\pm6.03$ & $\mathbf{99.92\pm0.06}$ 
        \\ \cmidrule{2-6}
        & \cellcolor{lightblue} \texttt{FedProx+PeWS} & \cellcolor{lightblue} $\mathbf{99.96\pm0.01}$ & \cellcolor{lightblue} $\mathbf{98.40\pm1.84}$ & \cellcolor{lightblue} $\mathbf{90.40\pm3.91}$ & \cellcolor{lightblue} $\mathbf{99.92\pm0.07}$ 
        \\ \bottomrule 
        \end{tabular}
    }
    \label{table: convergence-table-fedprox}
\end{table} 

\subsection{Wall-clock time vs. Accuracy} 
\begin{figure}[H]
    \centering
    \includegraphics[width=\linewidth]{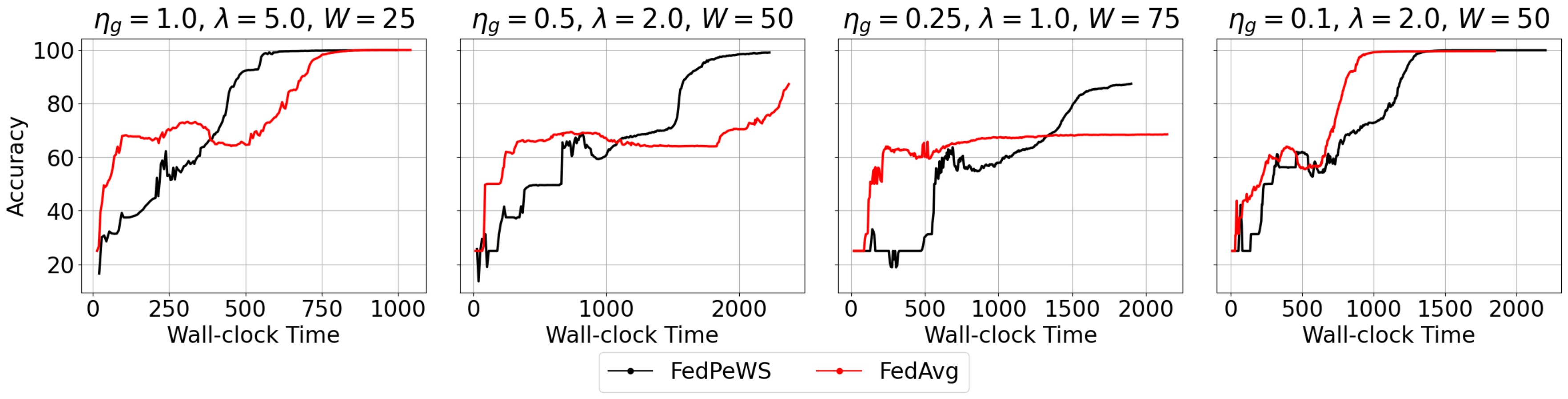}
    \caption{Wall-clock time vs. accuracy plot corresponding to Figure \ref{fig: synthetic-all} of the main paper. For better clarity, we selected a single run for each experiment at random.} 
    \label{fig: wall-clock}
\end{figure}

Figure \ref{fig: wall-clock} illustrates the wall-clock time versus accuracy results, which correspond to Figure \ref{fig: synthetic-all} in the main paper. From this comparison, \texttt{FedPeWS} demonstrates a slightly improved performance over FedAvg in terms of wall-clock time in two of the four scenarios. However, it underperforms slightly in the remaining two scenarios, with only a marginal increase in time. This variance is attributed to the alternation between training masks and weights during the warmup phase, impacting the time efficiency. 

\subsection{\texttt{FedPeWS-Fixed}. Mask Length Study} 
In this section, we explore the impact of mask length on the performance of the \texttt{FedPeWS-Fixed} method with parameter $W=120\ (\tau=0.4)$ using the CIFAR-MNIST dataset. We examine two scenarios for splitting the network into two subnetworks:

\begin{enumerate}
    \item $|m_1| < |m_2|$: 75\% of the mask is assigned to Participant 2, 25\% to Participant 1. 
    \item $|m_1| = |m_2|$: equal sized masks are assigned to each participant. 
\end{enumerate}

\begin{figure}[h] 
    \centering
    \includegraphics[width=0.8\linewidth]{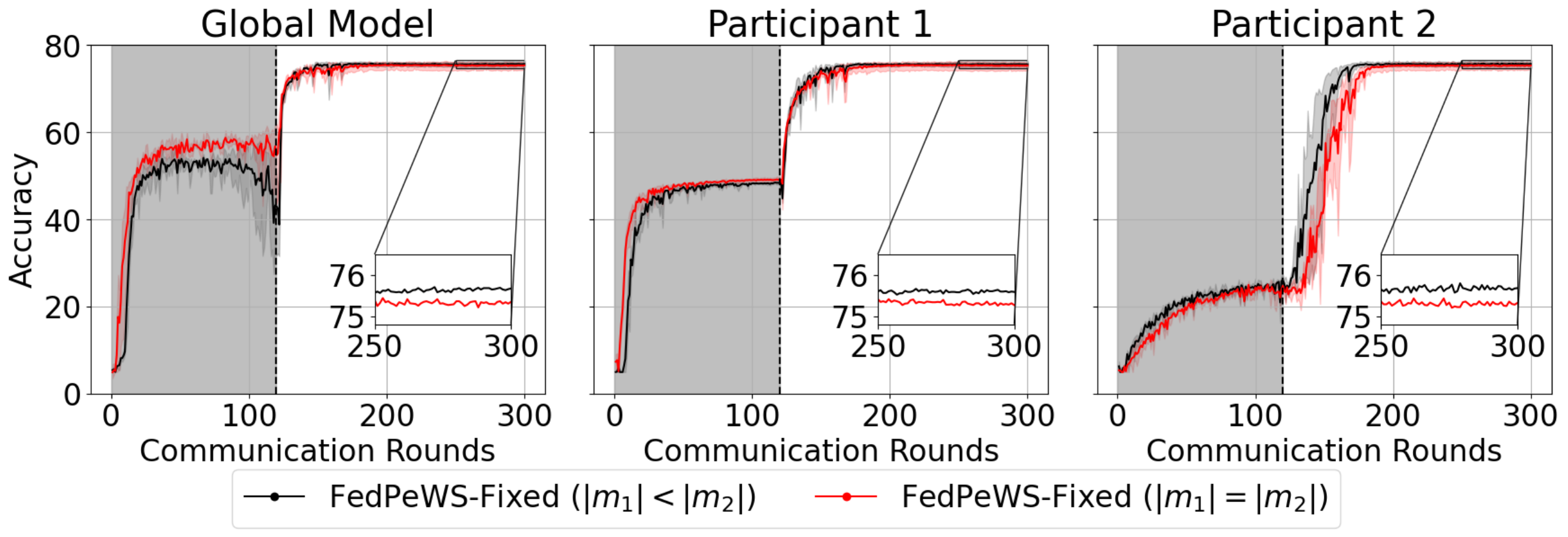} 
    \caption{Mask length study using \texttt{FedPeWS-Fixed} method on CIFAR-MNIST dataset. }
    \label{fig: mask-length-study}
\end{figure}

Figure \ref{fig: mask-length-study} displays the validation accuracy over $T=300$ communication rounds for both scenarios. The leftmost plot shows the accuracy of the global model, while the middle and rightmost plots show the accuracy for each of the participants. Both mask length scenarios converge to comparable accuracy levels, with a marginal difference of $0.5\%$ higher accuracy in the scenario where $|m_1| < |m_2|$. This is likely due to the larger mask size, which aids in learning the more complex CIFAR-10 dataset held by Participant 2. 

\subsection{Sensitivity Analysis} 
In this section, we detail the sensitivity analysis of the $\lambda$ and $\tau$ parameters conducted on the CIFAR-MNIST dataset (Table \ref{tab: cifar-mnist-ablation}) and the \{Path-OCT-Tissue\}MNIST dataset (Table \ref{tab: pot-ablation}). This analysis particularly includes the standard deviation of the accuracy achieved by the tested algorithms after $T$ communication rounds and over three independent evaluations. Results with the best performance are highlighted in green. 

\begin{table}[b]
    \centering
    \caption{Ablation study of the parameters $\lambda$ and $\tau$ on the CIFAR-MNIST dataset with $N=2$ participants over three independent runs. The first column ($\tau=0.0$) corresponds to the FedAvg algorithm. The last row $(-)$ presents results for the \texttt{FedPeWS-Fixed} algorithm.} 
    \resizebox{\linewidth}{!}{ 
    \begin{tabular}{r|cccccc} 
        \toprule
        & \multicolumn{6}{c}{\cellcolor{lightgray}\textbf{Proportion of warmup rounds $\tau = W/T$}} \\ \cmidrule{2-7} 
        $\mathbf{\lambda} (\downarrow) , \tau (\rightarrow)$ & $\tau=0.0$ (FedAvg) & $\tau=0.2$ & $\tau=0.4$ & $\tau=0.5$ & $\tau=0.6$ & $\tau=0.8$ \\ \midrule 
        0.0 & \multirow{12}{*}{$71.23 \pm 0.71$} & $68.01 \pm 0.88$ & $66.01\pm0.48$ & $65.96\pm1.03$ & $65.40\pm1.95$ & $65.77\pm0.41$ \\ \cmidrule{3-7}
        0.1 & & $68.77\pm1.09$ & $70.39\pm1.00$ & $70.61\pm1.17$ & $69.48\pm0.37$ & $70.36\pm2.06$ \\
        0.3 & & $70.86\pm0.23$ & $73.00\pm0.65$ & $73.91\pm0.71$ & $73.26\pm0.46$ & $73.02\pm1.05$ \\
        0.5 & & $71.43\pm0.56$ & $74.17\pm0.91$ & $73.84\pm0.12$ & $73.66\pm0.88$ & $75.05\pm0.45$ \\
        1.0 & & $72.26\pm0.54$ & $74.46\pm0.44$ & $74.54\pm0.76$ & $74.91\pm0.42$ & $73.81\pm0.87$ \\
        2.0 & & $72.61\pm0.79$ & $73.68\pm0.17$ & \cellcolor{lightgreen} $\mathbf{75.35\pm0.50}$ & $74.76\pm0.54$ & $74.46\pm0.56$ \\
        5.0 & & $72.60\pm0.45$ & \cellcolor{lightgreen} $\mathbf{75.22\pm0.33}$ & \cellcolor{lightgreen} $\mathbf{75.00\pm0.74}$ & $75.01\pm0.71$ & $73.96\pm1.60$ \\
        10.0 & & $72.29\pm0.48$ & \cellcolor{lightgreen} $\mathbf{74.97\pm0.65}$ & $74.31\pm0.95$ & $74.03\pm0.30$ & $71.91\pm2.69$ \\
        100.0 & & $71.64\pm0.47$ & $72.92\pm0.39$ & $73.96\pm0.65$ & $73.13\pm0.73$ & $72.43\pm3.55$ \\
        1000.0 & & $71.58\pm0.53$ & $73.18\pm0.73$ & $73.32\pm1.70$ & $73.87\pm1.16$ & $72.52\pm1.92$ \\ \cmidrule{3-7} 
        $-$ & & $72.72\pm0.44$ & \cellcolor{lightgreen} $\mathbf{75.22\pm0.19}$ & \cellcolor{lightgreen} $\mathbf{75.05\pm0.42}$ & $72.51\pm3.89$ & $73.77\pm0.20$ \\ 
        \bottomrule
    \end{tabular}
    }
    \label{tab: cifar-mnist-ablation}
\end{table} 
\begin{table}[t]
    \centering
    \caption{Ablation study of the parameters $\lambda$ and $\tau$ on the combination of \{Path-OCT-Tissue\}MNIST datasets with $N=3$ participants over three independent runs. The first column ($\tau=0.0$) corresponds to the FedAvg algorithm. The last row $(-)$ presents results for the \texttt{FedPeWS-Fixed} algorithm.} 
    \resizebox{\linewidth}{!}{ 
    \begin{tabular}{r|cccccc} 
        \toprule
        & \multicolumn{6}{c}{\cellcolor{lightgray}\textbf{Proportion of warmup rounds $\tau = W/T$}} \\ \cmidrule{2-7} 
        ${\lambda} (\downarrow) , \tau (\rightarrow)$ & $\tau=0.0$ (FedAvg) & $\tau=0.2$ & $\tau=0.4$ & $\tau=0.5$ & $\tau=0.6$ & $\tau=0.8$ \\ \midrule 
        0.0 & \multirow{12}{*}{$52.25\pm0.57$} & $53.04\pm2.08$ & $50.35\pm0.61$ & $47.83\pm1.04$ & $47.89\pm1.17$ & $46.80\pm1.76$ \\ \cmidrule{3-7}
        0.1 & & $52.52\pm1.79$ & $51.72\pm0.63$ & $49.83\pm1.83$ & $48.83\pm2.51$ & $47.50\pm1.07$ \\
        0.3 & & $52.74\pm1.75$ & $54.36\pm0.80$ & $51.07\pm0.25$ & $50.94\pm2.22$ & $50.62\pm1.91$ \\
        0.5 & & $54.30\pm2.08$ & $54.42\pm1.43$ & $52.02\pm2.53$ & $51.43\pm1.60$ & $48.97\pm0.80$ \\
        1.0 & & \cellcolor{lightgreen} $\mathbf{54.89\pm0.72}$ & $53.59\pm1.40$ & $51.49\pm1.19$ & $50.29\pm2.70$ & $52.21\pm1.02$ \\
        2.0 & & \cellcolor{lightgreen} $\mathbf{54.75\pm1.12}$ & \cellcolor{lightgreen} $\mathbf{54.45\pm1.78}$ & $53.02\pm0.45$ & $52.26\pm1.81$ & $52.49\pm0.49$ \\ 
        5.0 & & \cellcolor{lightgreen} $\mathbf{54.91\pm0.91}$ & \cellcolor{lightgreen} $\mathbf{54.99\pm0.90}$ & $52.41\pm0.45$ & $52.95\pm1.49$ & $52.53\pm1.35$ \\ 
        10.0 & & \cellcolor{lightgreen} $\mathbf{55.12\pm1.16}$ & \cellcolor{lightgreen} $\mathbf{55.03\pm1.39}$ & $52.85\pm0.95$ & $50.82\pm3.03$ & $52.27\pm1.11$ \\ 
        100.0 & & $54.22\pm1.74$ & $52.31\pm2.55$ & $52.10\pm0.87$ & $49.52\pm4.34$ & $51.46\pm2.92$ \\ 
        1000.0 & & $53.45\pm1.65$ & $53.82\pm2.16$ & $51.16\pm1.92$ & $51.30\pm2.02$ & $51.19\pm1.05$ \\ \cmidrule{3-7} 
        $-$ & & $53.69\pm0.77$ & $51.78\pm0.44$ & $50.24\pm1.88$ & $51.12\pm0.72$ & $49.87\pm1.23$ \\ 
        \bottomrule
    \end{tabular}
    }
    \label{tab: pot-ablation}
\end{table} 

For the CIFAR-MNIST dataset, the preferred values of $\lambda$ that yield optimal outcomes range within $\{2.0, 5.0, 10.0\}$, and for $\tau$, the values are $\{0.4, 0.5\}$. A similar pattern is observed in the \{Path-OCT-Tissue\}MNIST dataset experiment, with a small difference, in which it shows a preference for fewer warmup rounds ($\tau \in \{0.2, 0.4\}$) and demonstrates optimal performance with the same set of $\lambda$ values. This consistency across different datasets indicates robustness in the parameter settings for achieving high accuracy.

\subsection{Larger number of participants} 

\subsubsection{$\mathbf{N=\{5, 10, 20\}}$ participants.} 
\begin{figure}[h]
    \centering
    \includegraphics[width=0.85\linewidth]{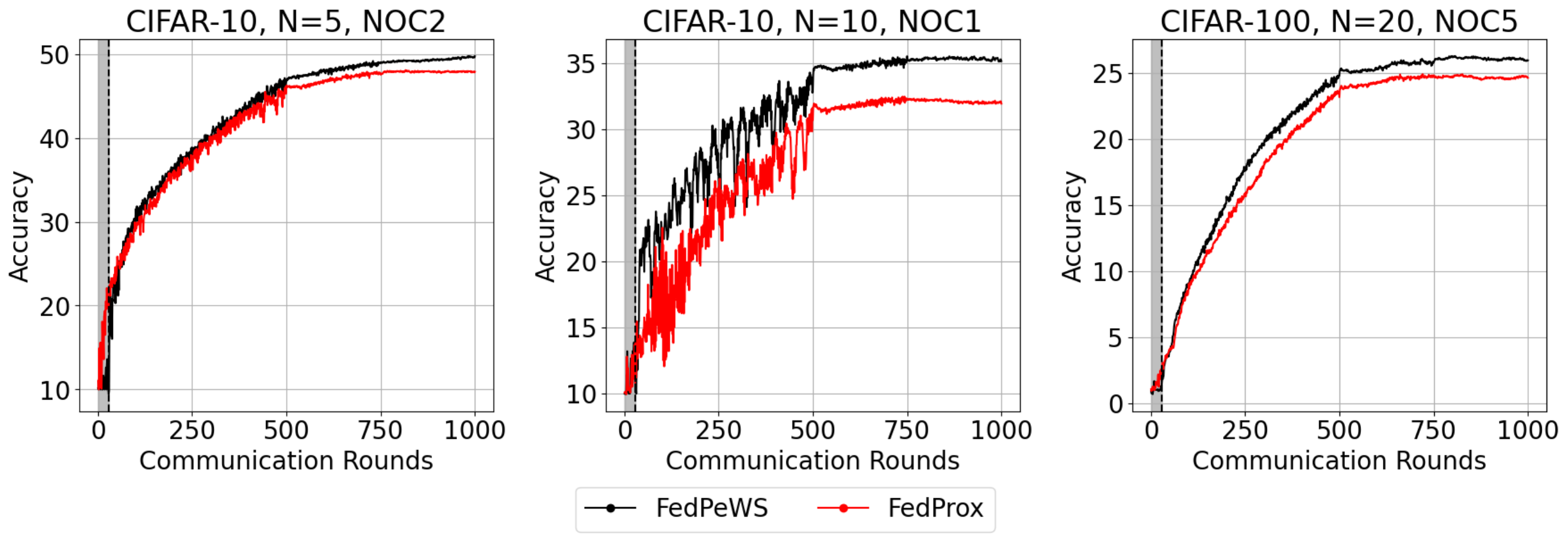}
    \caption{Visualization of global model performance with $N={5,10,20}$ participants on the CIFAR-10 and CIFAR-100 datasets. \textbf{\textit{Note:}} NOC refers to non-overlapping classes partition; e.g. for the CIFAR-100 experiment, we randomly select 5 classes out of 100 for each of the 20 participants without replacement.} 
    \label{fig: diff-num-of-participants}
\end{figure}

Figure \ref{fig: diff-num-of-participants} demonstrates the performance of our proposed algorithm on the CIFAR-10 and CIFAR-100 datasets \cite{krizhevsky2009learning} using a non-overlapping class splitting strategy, where an equal number of classes are sampled without replacement, totaling to the number of participants multiplied by the number of classes assigned to each participant. This represents a heterogeneous setting. For this comparison, we use FedProx with a scalar multiplier of $0.01$ as the baseline algorithm. The FedPeWS algorithm combines FedProx and our proposed method, referred to as FedProx+PeWS. As evident from the plots, our algorithm consistently outperforms the baseline, delivering a significant performance boost in heterogeneous scenarios and showcasing its effectiveness across various settings. 

\subsubsection{$\mathbf{N=200}$ participants.} 
\begin{figure}[h]
    \centering
    \includegraphics[width=0.6\linewidth]{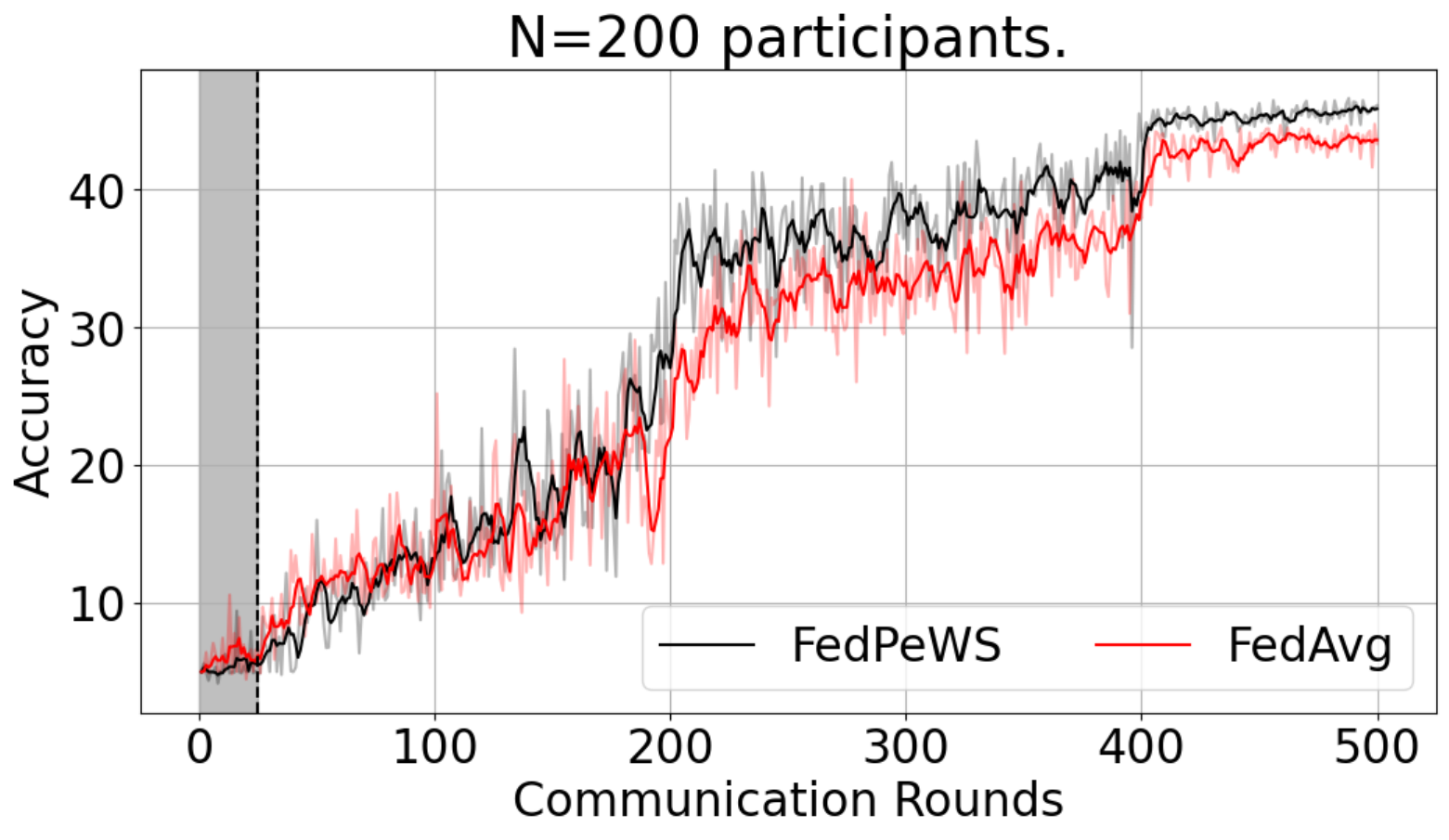}
    \caption{Visualization of global model performance with $N=200$ participants with a partial participation rate of $0.1$. Smoothing is applied as a running average with a window size of 5. A learning rate scheduler is implemented at rounds $200$ and $400$ with a learning rate decay factor of $0.1$.} 
    \label{fig: cifar-mnist-n200}
\end{figure}

Although our primary focus is on the cross-silo setting, we extend our study to include a large-scale scenario involving 200 participants on the CIFAR-MNIST dataset. We adopt a Dirichlet partition strategy with concentration parameter $\alpha=0.5$ and implement this scenario with a partial participation rate of $0.1$. The outcomes of this experiment, as depicted in Figure \ref{fig: cifar-mnist-n200}, indicate a superior performance compared to the conventional FedAvg algorithm, thereby further substantiating the validity and effectiveness of our proposed method. The parameters set for \texttt{FedPeWS} are: $W=25$ and $\lambda=0.5$. 

\section{Neuron Activations Study} 
In this section, we examine the extent to which neurons in each layer are activated. Our study uses the Synthetic-32K dataset and the \texttt{FedPeWS-Fixed} method (with parameter $W = 50$). The vertical dashed line ($W=50$) indicates the point at which participants switch to using full masks. 

\begin{figure}[h]
    \centering
    \includegraphics[width=\linewidth]{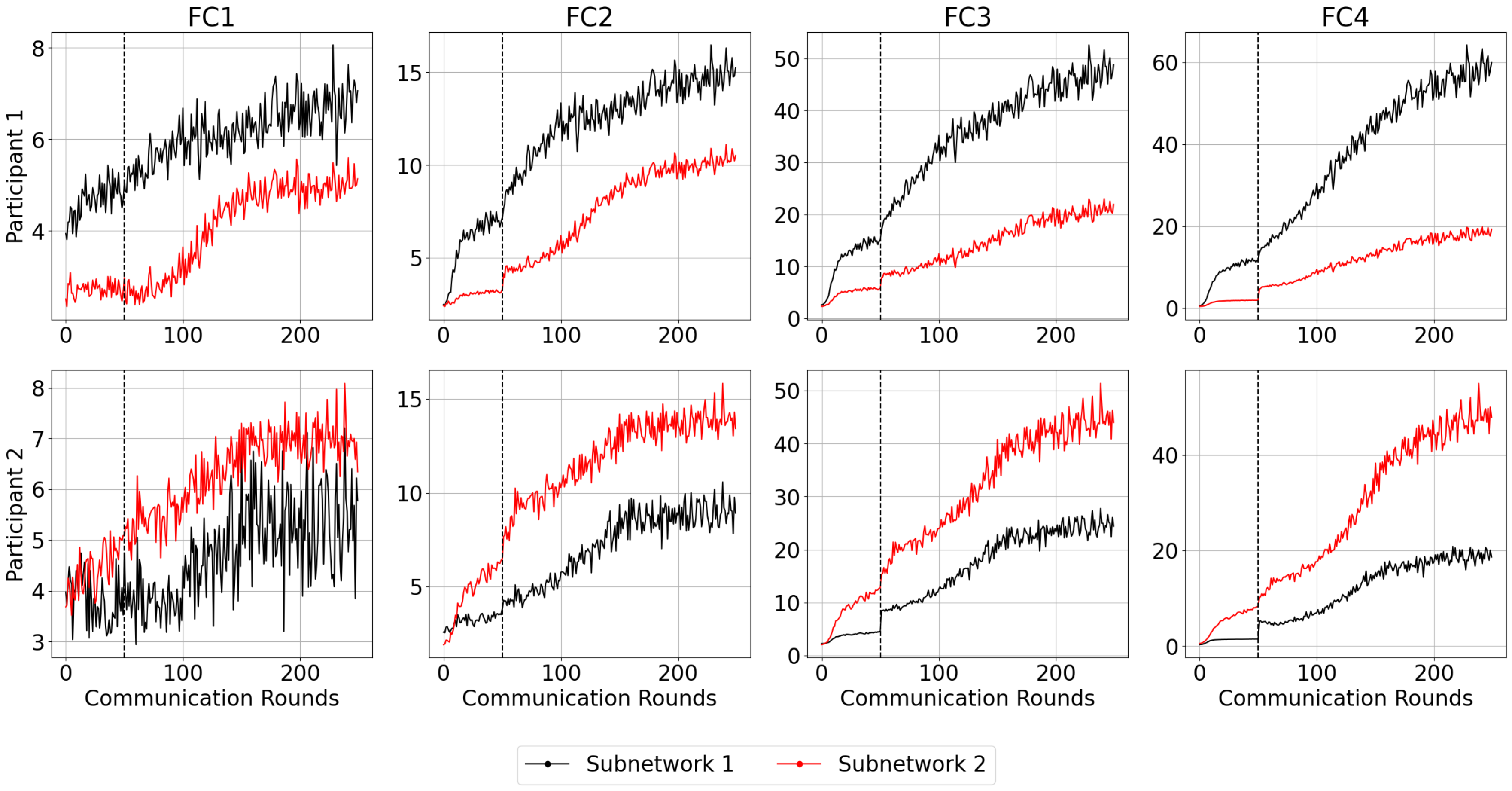}
    \caption{Neuron activation study on the Synthetic-32K dataset with a global learning rate $\eta_g=1.0$. The experiment uses the \texttt{FedPeWS-Fixed} method with parameter $W=50$, indicated by the vertical dashed line, marking the switch to full masks by each participant.} 
    \label{fig: neuron-activations-train}
\end{figure}

Figure \ref{fig: neuron-activations-train} displays the neuron activations, measured as the sum of activations over a batch of samples randomly selected from each participant's dataset, over $T=250$ communication rounds. The top row shows the outcomes for Participant 1, and the bottom row shows the activations for Participant 2. Each column corresponds to different fully connected layers (FC1 to FC4) in the network. Observations are as follows: 

\begin{enumerate}
    \item Before switching $(t \leq W)$: for Participant $i$, subnetwork $i$ shows higher activation patterns in all given FC layers, $i \in [1,2]$, while the other subnetwork exhibits a minimal activation. 
    \item After switching to full mask $(t > W)$: (i) there is a noticeable increase in activations for both participants upon switching to using full masks, (ii) Participant 1 with its originally initialized subnetwork 1, shows a substantial increase in activations compared to subnetwork 2 across all layers. The same pattern is observed for Participant 2 with subnetwork 2. 
\end{enumerate}

\begin{figure}[h] 
    \centering 
    \includegraphics[width=\linewidth]{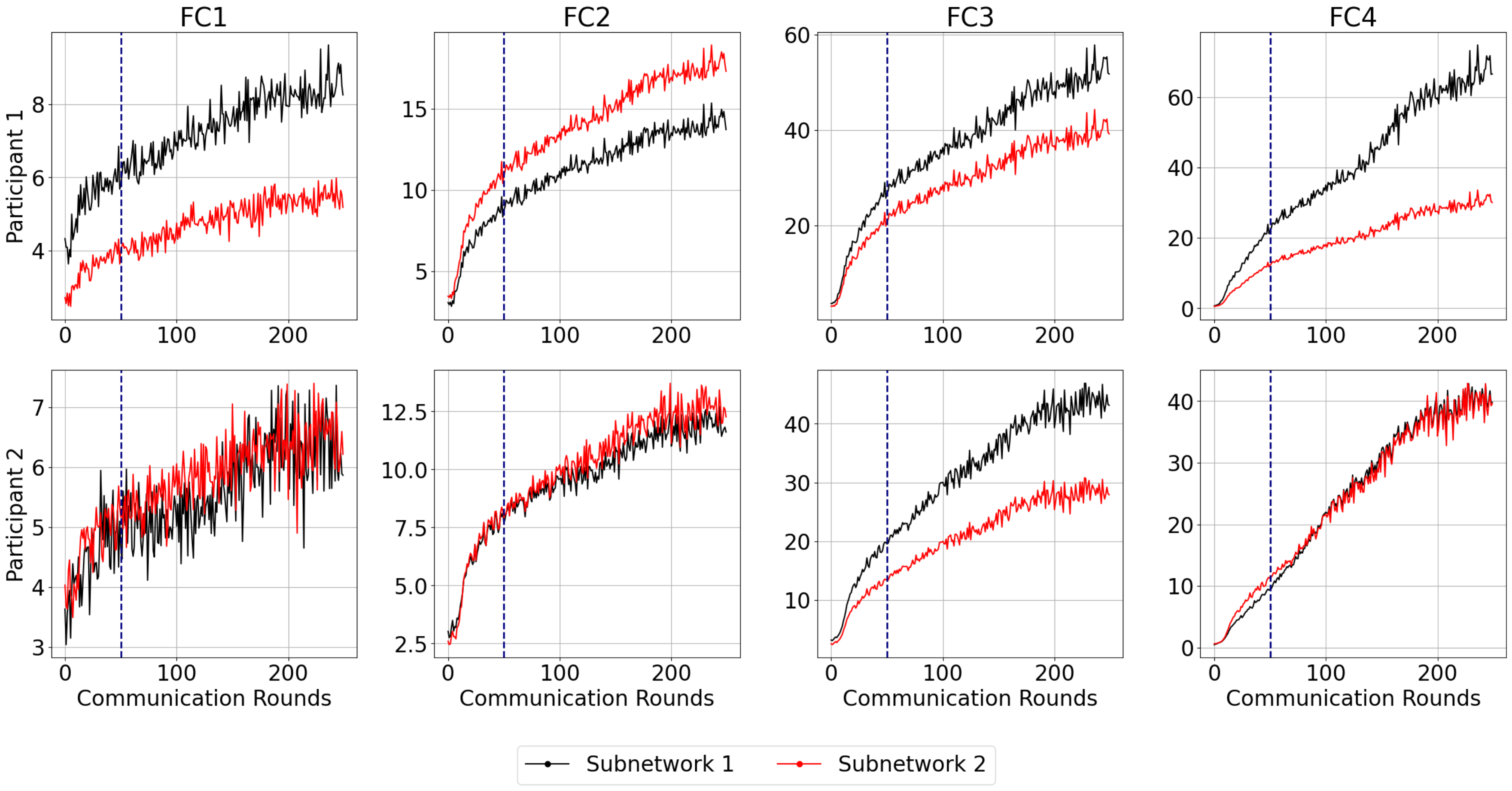} 
    \caption{Neuron activation study on the Synthetic-32K dataset with a global learning rate $\eta_g=1.0$ using FedAvg algorithm.} 
    \label{fig: neural-act-study-fedavg} 
\end{figure}

These findings suggest that the personalized warmup strategy helps the network learn which paths to follow when specific data points are fed into the network. This supports the superiority of our method and corroborates the claims made in the main paper. On the other hand, for the FedAvg algorithm (see Figure \ref{fig: neural-act-study-fedavg}), we observe the absence of a similar activation pattern. Data points from different participants activate overlapping regions of the model, which results in slower convergence.

\end{document}